%% file: main.tex
\documentclass[letterpaper,twocolumn,10pt]{article}
\usepackage{usenix-2020-09}

\input{macros}

\usepackage{microtype}
\usepackage{graphicx}
\usepackage{caption}
\usepackage{subcaption}
\usepackage{booktabs}
\usepackage{ifthen}
\usepackage{xcolor}
\usepackage{amsmath}
\usepackage{amssymb}
\usepackage{amsthm}
\usepackage{amsfonts}
\usepackage{tikz}
\usepackage{comment}
\usepackage{enumitem}
\usetikzlibrary{arrows.meta, positioning, shapes, calc}

\usepackage[most]{tcolorbox}
\tcbset{
  insightbox/.style={
    colback=blue!4!white,
    colframe=blue!40!black,
    boxrule=0.6pt,
    arc=2pt,
    left=6pt,
    right=6pt,
    top=6pt,
    bottom=6pt
  }
}

\makeatletter
\renewcommand{\sectionautorefname}{\S\@gobble}
\renewcommand{\subsectionautorefname}{\S\@gobble}
\renewcommand{\subsubsectionautorefname}{\S\@gobble}
\makeatother

\usepackage{xurl}
\usepackage{hyperref}

\usepackage{pifont}
\usepackage{multirow}
\definecolor{invgreen}{RGB}{34,139,34}
\definecolor{invred}{RGB}{178,34,34}

\newcommand{\cmark}{\textcolor{invgreen}{\ding{51}}}
\newcommand{\xmark}{\textcolor{invred}{\ding{55}}}

\begin{document}

\newboolean{publicversion}
\setboolean{publicversion}{true}
\ifthenelse{\boolean{publicversion}}{
    \newcommand{\todo}[1]{}
    \newcommand{\grumbler}[3]{}
    \newcommand{\ak}[1]{}
    \newcommand{\ap}[1]{}
    \newcommand{\raja}[1]{}
    \newcommand{\rr}[1]{}
    \newcommand{\simon}[1]{}
}
{
    \newcommand{\todo}[1]{\textcolor{red}{\bf #1}}
    \newcommand{\grumbler}[3]{\xspace\textcolor{#3}{\bf #1: #2}}
    \newcommand{\ak}[1]{\grumbler{Aditya}{#1}{violet}}
    \newcommand{\ap}[1]{\grumbler{Ashish}{#1}{red}}
    \newcommand{\raja}[1]{\grumbler{Raja}{#1}{purple}}
    \newcommand{\rr}[1]{\grumbler{Ram}{#1}{teal}}
    \newcommand{\simon}[1]{\grumbler{Simon}{#1}{red}}
}

\date{}

\title{\sysname: Enabling Determinism in LLM Inference with Verified Speculation}

\author{
{\rm Raja Gond$^{\ddagger}$, Aditya K Kamath$^{\dagger}$, Ramachandran Ramjee$^{\ddagger}$, and Ashish Panwar$^{\ddagger}$}
\\[1ex]
$^{\ddagger}$Microsoft Research \quad
$^{\dagger}$University of Washington
}

\maketitle

\input{Sections/0_Abstract}

\input{Sections/1_Introduction}

\input{Sections/2_Background}

\input{Sections/3_Observations}

\input{Sections/4_Design}
\input{Sections/5_Evaluation}
\input{Sections/6_RelatedWork}
\input{Sections/7_Conclusion}

\bibliographystyle{plain}
\bibliography{references}

\end{document}

%% file: macros.tex
\usepackage{xspace}
\newcommand{\sysname}{\textsc{LLM-42}\xspace}
\newcommand{\vllm}{\textsc{vLLM}\xspace}
\newcommand{\sglang}{SGLang\xspace}
\newcommand{\sglangdet}{SGLang-Deterministic\xspace}
\newcommand{\sglangnondet}{SGLang-Non-Deterministic\xspace}

\newcommand{\llama}{Llama-3.1-8B-Instruct\xspace}

\newcommand{\sharegpt}{ShareGPT\xspace}

\newcommand{\running}[1]{\noindent \textbf{#1:}}

%% file: Sections/0_Abstract.tex
\begin{abstract}

In LLM inference, the same prompt may yield different outputs across different runs. At the system level, this non-determinism arises from floating-point non-associativity combined with dynamic batching and GPU kernels whose reduction orders vary with batch size. A straightforward way to eliminate non-determinism is to disable dynamic batching during inference, but doing so severely degrades throughput. Another approach is to make kernels batch-invariant; however, this tightly couples determinism to kernel design, requiring new implementations. This coupling also imposes fixed runtime overheads, regardless of how much of the workload actually requires determinism.

Inspired by ideas from speculative decoding, we present \sysname{}\footnote{Many developers unknowingly set 42 as a random state variable (seed) to ensure reproducibility, a reference to \textit{The Hitchhiker's Guide to the Galaxy}. The number itself is arbitrary, but widely regarded as a playful tradition.}—a scheduling-based approach to enable determinism in LLM inference. Our key observation is that if a sequence is in a consistent state, the next emitted token is likely to be consistent even with dynamic batching. Moreover, most GPU kernels use shape-consistent reductions. Leveraging these insights, \sysname decodes tokens using a non-deterministic fast path and enforces determinism via a lightweight verify–rollback loop. The verifier replays candidate tokens under a fixed-shape reduction schedule, commits those that are guaranteed to be consistent across runs, and rolls back those violating determinism. \sysname mostly re-uses existing kernels unchanged and incurs overhead only in proportion to the traffic that requires determinism.

\end{abstract}

%% file: Sections/1_Introduction.tex
\section{Introduction} \label{sec:intro}

LLMs are becoming increasingly more powerful~\cite{openai2022gpt4techreport,kaplan2020scalinglaws}. However, a key challenge many users usually face with LLMs is their non-determinism~\cite{he2025nondeterminism, atil2024nondeterminism, yuan2025fp32death, song2024greedy}: the same model can produce different outputs across different runs of a given prompt, even with identical decoding parameters and hardware. Enabling determinism in LLM inference (aka deterministic inference) has gained significant traction recently for multiple reasons. It helps developers isolate subtle implementation bugs that arise only under specific batching choices; improves reward stability in reinforcement-learning training~\cite{zhang2025-deterministic-tp}; is essential for integration testing in large-scale systems. Moreover, determinism underpins scientific reproducibility~\cite{song2024greedy} and enables traceability~\cite{rainbird2025deterministic}. Consequently, users have increasingly requested support for deterministic inference in LLM serving engines~\cite{Charlie2025, Anadkat2025consistent}.

He et al.~\cite{he2025nondeterminism} showed that non-determinism in LLM inference at the system-level stems from the non-associativity of floating-point arithmetic\footnote{LLM outputs can also vary due to differences in sampling strategies (e.g., top-k, top-p, or temperature). Our goal is to ensure  that output is deterministic for fixed sampling hyper-parameters; different hyper-parameters can result in different output and this behavior is intentional.}. This effect manifests in practice because most core LLM operators—including matrix multiplications, attention, and normalization—rely on reduction operations, and GPU kernels for these operators choose different reduction schedules to maximize performance at different batch sizes.
The same study also proposed \emph{batch-invariant computation} as a means to eliminate non-determinism. In this approach, a kernel processes each input token using a single, universal reduction strategy independent of batching. Popular LLM serving systems such as \vllm and \sglang have recently adopted this approach~\cite{SGLangTeam2025,vllm-batch-invariant-2025}.

While batch-invariant computation guarantees determinism, we find that this approach is fundamentally over-constrained. Enforcing a fixed reduction strategy for every token—regardless of model phase or batch geometry—strips GPU kernels of the very parallelism they are built to exploit. For example, the batch-invariant GEMM kernels provided by He et al. do not use the split-K strategy that is otherwise commonly used to accelerate GEMMs at low batch sizes~\cite{tritonfusedkernel-splitk-meta,nvidia_cutlass_blog}. Furthermore, most GPU kernels are not batch-invariant to begin with, so insisting on batch-invariant execution effectively demands new kernel implementations, increasing engineering and maintenance overhead. Finally, batch-invariant execution makes determinism the default for all requests, even when determinism is undesirable or even harmful~\cite{det-inf-kills}.

Our observations suggest that determinism can be enabled with a simpler approach. \textbf{(O1)} Token-level inconsistencies are rare: as long as a sequence remains in a consistent state, the next emitted token is mostly identical across runs; sequence-level divergence arises mainly from autoregressive decoding after the first inconsistent token.  \textbf{(O2)} Most GPU kernels already use shape-consistent reduction schedules: they apply the same reduction strategy on all inputs of a given shape, but potentially different reduction strategies on inputs of different shapes. \textbf{(O3)} Determinism requires only position-consistent reductions: a particular token position must use the same reduction schedule across runs, but different positions within or across sequences can use different reduction schedules. \textbf{(O4)} Real-world LLM systems require determinism only for select tasks (e.g., evaluation, auditing, continuous-integration pipelines), while creative workloads benefit from the stochasticity of LLMs.

\begin{figure}[t!]
    \centering
    \includegraphics[width=0.75\linewidth]{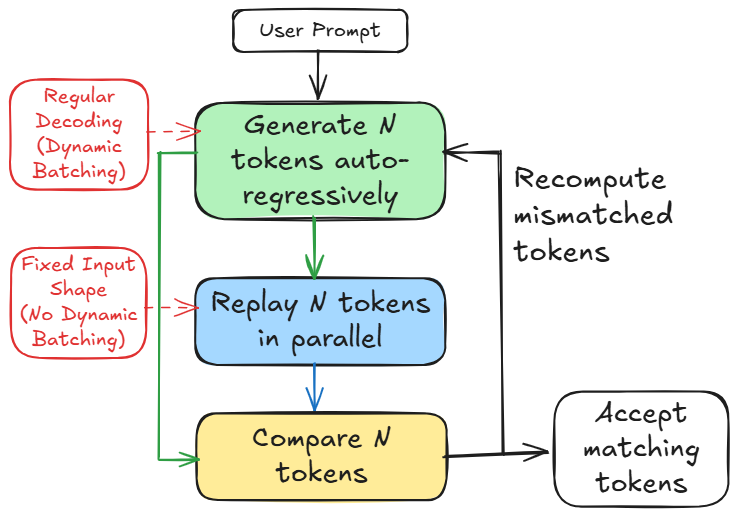}
    \caption{Overview of \sysname.}
    \label{fig:introduction:banner}
\end{figure}

Based on these observations, we introduce \sysname{}, a scheduling-based approach to deterministic LLM inference inspired by speculative decoding. In speculative decoding~\cite{specdecoding-icml2023,chen2023accelerating,xia2023speculative,specinfer-2024}, a fast path produces multiple candidate tokens while a verifier validates their correctness. We observe that the same structure can be repurposed to enable determinism (see~\autoref{fig:introduction:banner}): fast path can optimize for high throughput token generation while a verifier can enforce determinism. \autoref{tab:speculative_vs_llm42} highlights key differences between conventional speculative decoding and our approach.

\sysname{} employs a \emph{decode–verify–rollback} protocol that decouples regular decoding from determinism enforcement. It generates candidate output tokens using standard fast-path autoregressive decoding whose output is largely—but not provably—consistent across runs (O1). Determinism is guaranteed by a verifier that periodically replays a fixed-size window of recently generated tokens. Because the input shape is fixed during verification, replayed tokens follow a consistent reduction order (O2) and serve as a deterministic reference execution. Tokens are released to the user only after verification. When the verifier detects a mismatch, \sysname{} rolls the sequence back to the last matching token and resumes decoding from this known-consistent state. In general, two factors make this approach practical: (1) verification is low cost i.e., like prefill, it is typically 1-2 orders of magnitude more efficient than the decode phase, and (2) rollbacks are infrequent: more than half of the requests complete without any rollback, and only a small fraction require multiple rollbacks.

\begin{table}[t]
\centering
\scriptsize
\setlength{\tabcolsep}{6pt}
\renewcommand{\arraystretch}{1.35}
\begin{tabular}{p{0.45\columnwidth} p{0.45\columnwidth}}
\toprule
\textbf{Speculative Decoding} &
\textbf{\sysname} \\
\midrule
Fast path generates tokens using some form of approximation &
No approximation; only floating-point rounding errors \\[2pt]

Low acceptance rate and hence limited speculation depth (2-8 tokens) &
High acceptance rate and hence longer speculation (32-64 tokens) \\[2pt]

Typically uses separate draft and target models &
Uses the same model for decoding and verification \\
\bottomrule
\end{tabular}
\caption{Speculative decoding vs. \sysname.}
\label{tab:speculative_vs_llm42}
\end{table}

By decoupling determinism from token generation, \sysname{} enables determinism to be enforced selectively, preserving the natural variability and creativity of LLM outputs where appropriate. This separation also helps performance: the fast path can use whatever batch sizes and reduction schedules are most efficient, prefill computation can follow different reduction strategy than decode (O3) and its execution need not be verified (in our design, prefill is deterministic by construction), and finally, verification can be skipped entirely for requests that do not need determinism (O4).

The efficiency of our approach critically depends on the size of verification window i.e., number of tokens verified together. Smaller windows incur high verification overhead since their computation is largely memory-bound but require less recomputation on verification failures. In contrast, larger windows incur lower verification cost due to being compute-bound, but increase recomputation cost by triggering longer rollbacks on mismatches. To balance this trade-off, we introduce \emph{grouped verification}: instead of verifying a large window of a single request, we verify smaller fixed-size windows of multiple requests together. This design preserves the low rollback cost of small windows while amortizing verification overhead. Overall, we make the following contributions:

\begin{itemize}
\item We present the first systematic analysis of batch-invariant computation to highlight the performance and engineering cost associated with this approach.

\item We present an alternate approach \sysname to enable determinism in LLM inference.

\item We implement \sysname on top of \sglang and show that its overhead is proportional to fraction of traffic that requires determinism; it retains near-peak performance when deterministic traffic is low, whereas \sglang incurs high overhead of up to 56\% in deterministic mode. Our source code will be available at \url{https://github.com/microsoft/llm-42}.

\end{itemize}

%% file: Sections/2_Background.tex
\section{Background and Motivation}

\begin{figure}[t!]
    \centering
    \includegraphics[width=0.98\linewidth]{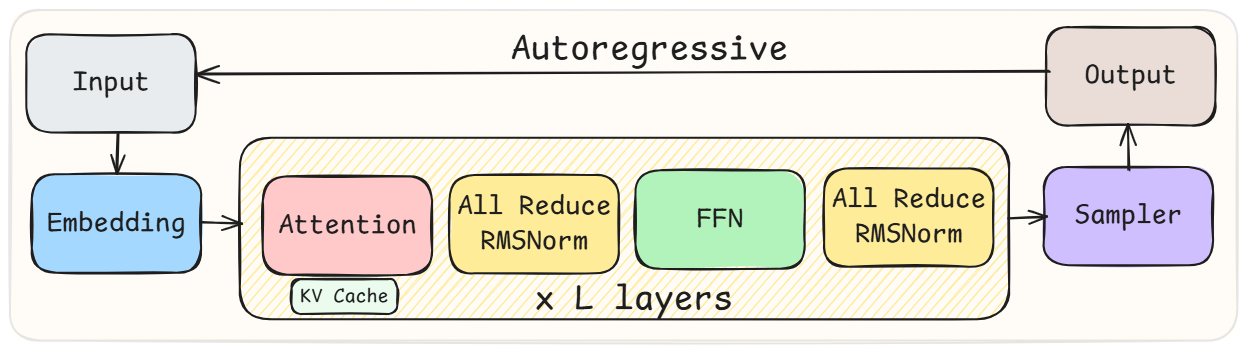}
    \caption{High-level architecture of LLMs.}
    \label{fig:background:arch-llms}
\end{figure}

This section introduces LLMs, explains the source of non-determinism in LLM inference and quantifies the cost of batch-invariant computation based deterministic inference.

\subsection{Large Language Models}

Transformer-based LLMs compose a stack of identical decoder blocks, each containing a self-attention module, a feed-forward network (FFN), normalization layers and communication primitives (\autoref{fig:background:arch-llms}). The hidden state of every token flows through these blocks sequentially, but within each block the computation is highly parallel: attention performs matrix multiplications over the key–value (KV) cache, FFNs apply two large GEMMs surrounding a nonlinearity, and normalization applies a per-token reduction over the hidden dimension.

Inference happens in two distinct phases namely prefill and decode. Prefill processes prompt tokens in parallel, generating KV cache for all input tokens. This phase is dominated by large  parallel computation. Decode is autoregressive: each step consumes the most recent token, updates the KV cache with a single new key and value, and produces the next output token. Decode is therefore sequential within a sequence but parallel across other requests in the batch.

\subsection{Non-determinism in LLM Inference}

In finite precision, arithmetic operations such as accumulation are non-associative, meaning that $(a+b)+c \neq a+(b+c)$. Non-determinism in LLM inference stems from this non-associativity when combined with \textit{dynamic batching}, a standard technique to achieve high throughput inference. With dynamic batching, the same request may be co-located with different sets of requests across different runs, resulting in varying batch sizes. Further, GPU kernels adapt their parallelization---and consequently their reduction---strategies based on the input sizes. As a result, the same logical operation can be evaluated with different floating-point accumulation orders depending on the batch it appears in, leading to inconsistent numerical results.

\begin{figure}[t!]
    \centering
    \begin{subfigure}[b]{0.42\linewidth}
        \centering
        \includegraphics[width=0.9\linewidth]{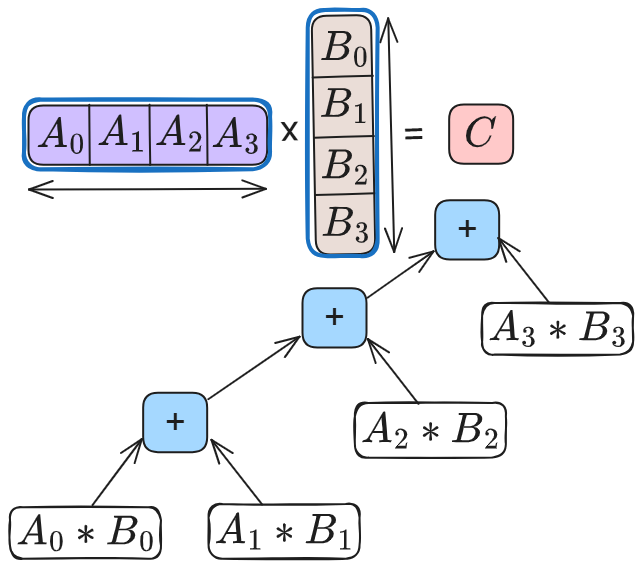}
        \caption{Without split-K.}
        \label{subfig:compute}
    \end{subfigure}
    \hfill
    \begin{subfigure}[b]{0.54\linewidth}
        \centering
        \includegraphics[width=0.9\linewidth]{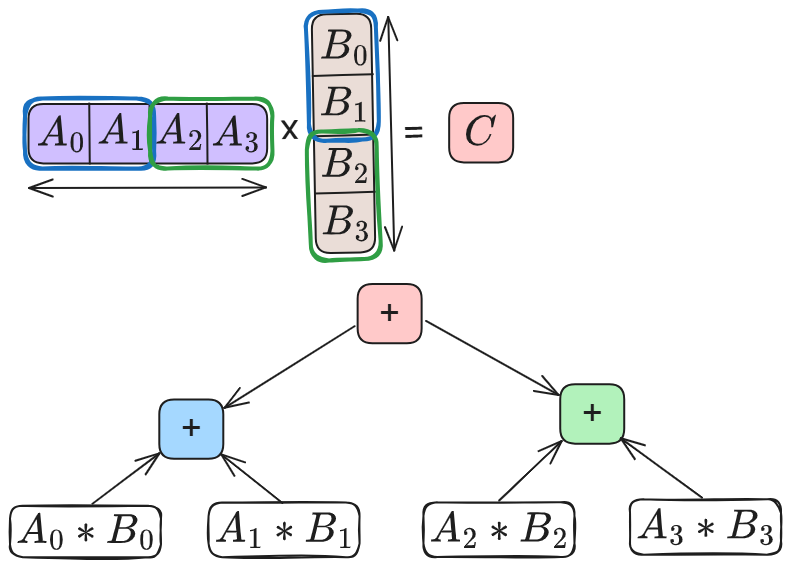}
        \caption{With split-K.}
        \label{subfig:parallelism}
    \end{subfigure}
    \caption{GEMM kernels compute dot products using standard accumulation or split-K parallelization. While split-K increases parallelism, it alters the reduction tree based on K.}
    \label{fig:computation_methods}
\end{figure}

Reductions are common in LLM inference, appearing in matrix multiplications (GEMMs), attention, normalization and collective communication such as AllReduce.  GEMM is the most common and time consuming operator. High-performance GEMM implementations on GPUs use hierarchical tiling and parallel reductions to improve occupancy and memory reuse. A common optimization is split-K parallelism, where the reduction dimension is partitioned across multiple thread blocks. Each block computes a partial result, which is then combined via an additional reduction step. Whether split-K is used—and how many splits are chosen—depends on the shape on input matrices.  These choices directly change the reduction tree as shown in~\autoref{fig:computation_methods}, producing different numerical results for a given token across runs. Similarly, attention kernels split work across the key–value dimension to increase SM utilization, followed by a reduction to combine partial results. Normalization operators reduce across feature dimensions. As inference progresses, batch-dependent reduction choices in these operators introduce small numerical drifts that propagate across kernels, layers and decoding steps, eventually influencing output tokens even when the sampling hyper-parameters are fixed.

\subsection{Defeating Non-determinism}

He et al. at Thinking Machines Lab recently introduced a new approach named batch-invariant computation to enable deterministic LLM inference~\cite{he2025nondeterminism}. In this approach, a GPU kernel is constrained to use a single, universal reduction strategy for all tokens, eliminating batch-dependent reductions. Both \sglang and \vllm use this approach~\cite{SGLangTeam2025,vllm-batch-invariance-2025,vllm-batch-invariant-2025}: these systems either deploy the batch-invariant kernels provided by He et al.~\cite{tmops2025} or implement new kernels~\cite{batch-inv-inf-vllm,det-infer-batch-inv-ops-sglang}.

\begin{figure}[t]
  \centering
  \begin{subfigure}{0.65\columnwidth}
    \centering
    \includegraphics[width=\columnwidth]{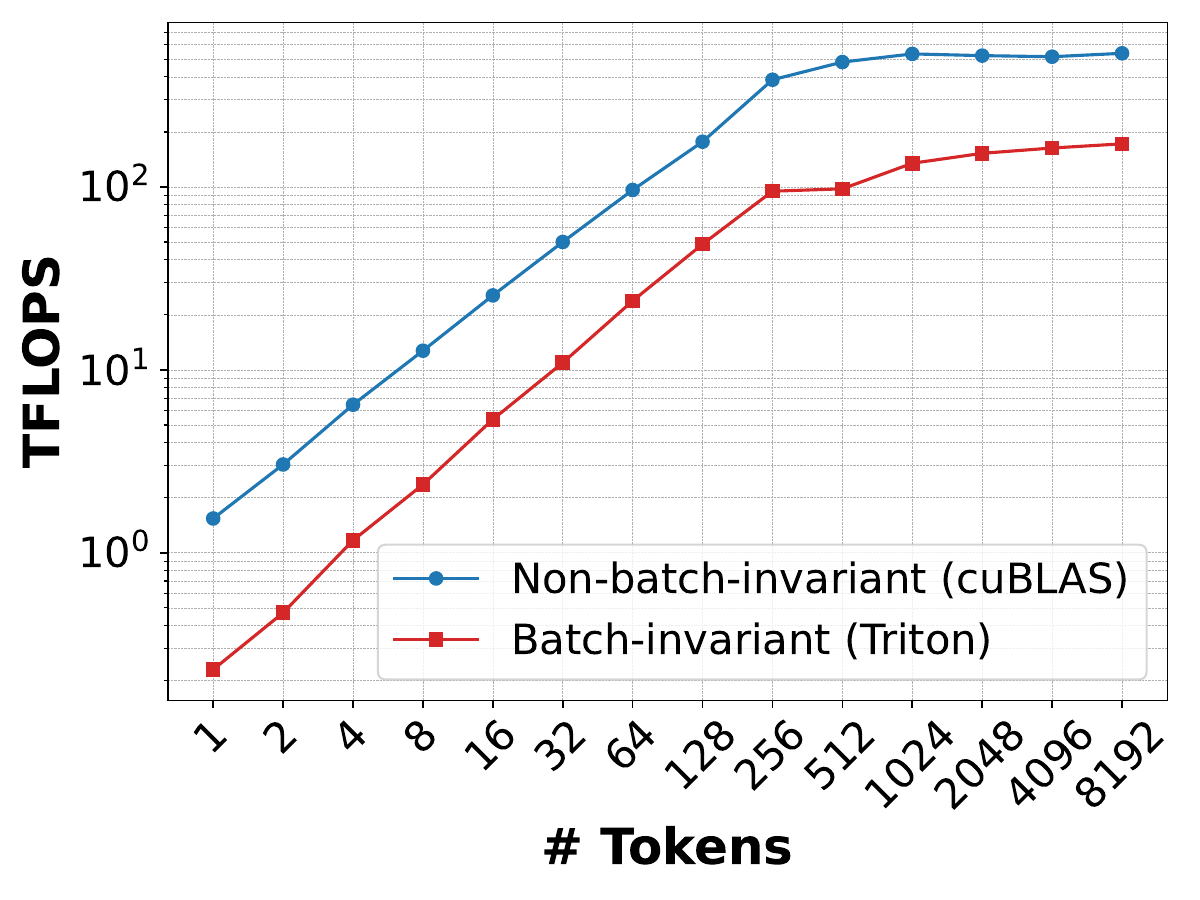}
    \caption{GEMM}
    \label{fig:motivation:prepej}
  \end{subfigure}
  \hfill
  \begin{subfigure}{0.65\columnwidth}
    \centering
    \includegraphics[width=\columnwidth]
    {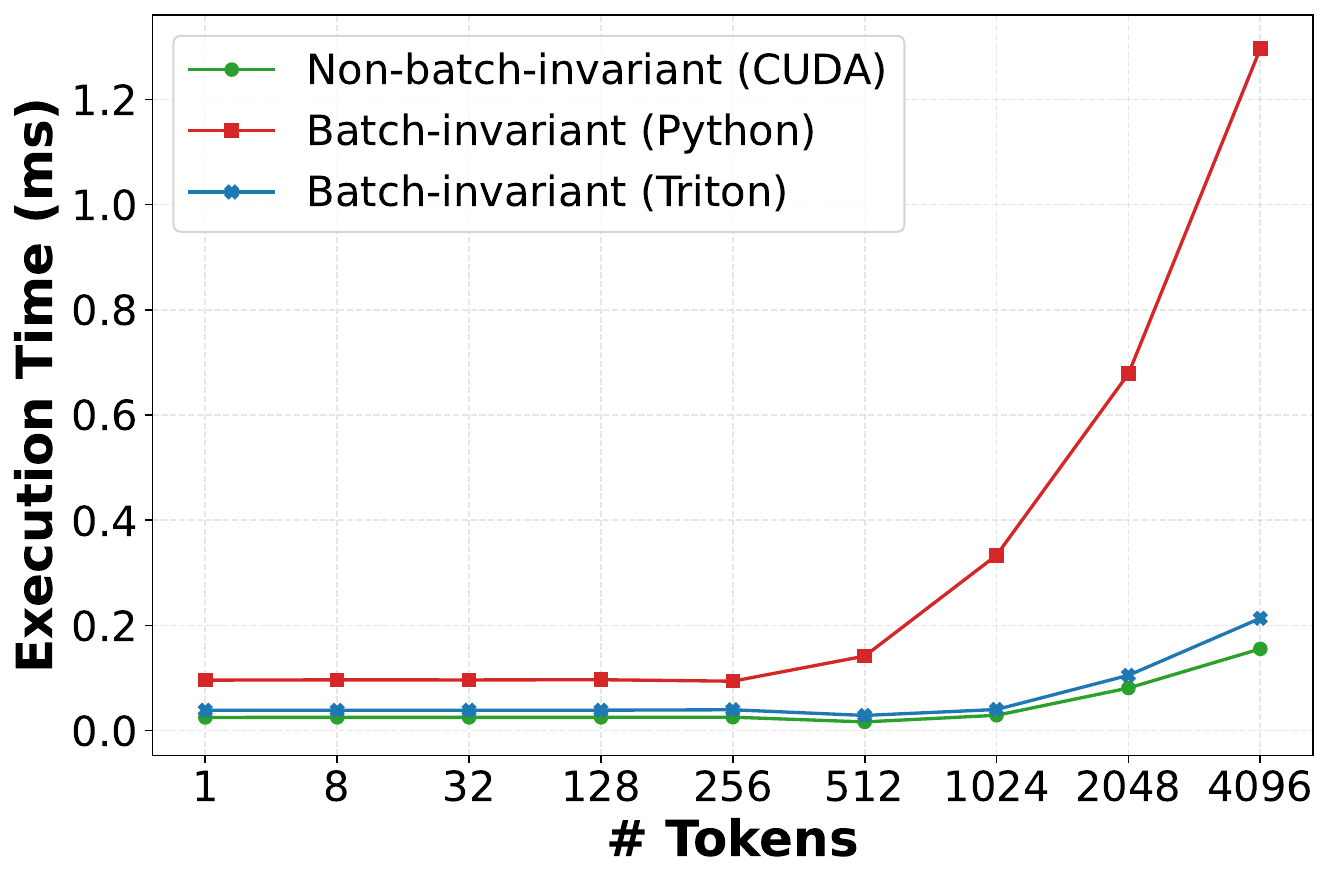}
    \caption{RMSNorm}
    \label{fig:motivation:rmsnorm_impl}
  \end{subfigure}
  \caption{Performance comparison between batch-invariant vs. non-batch-invariant kernels.}
  \label{fig:motivation:operator-perf-combined}
\end{figure}

While batch-invariant computation eliminates non-determinism, we find that it is a poor fit for real LLM serving systems. By enforcing a universal reduction strategy across all executions, it couples determinism to kernel design and sacrifices performance opportunities. It also turns determinism into a fixed tax paid by every request: dynamic batching aggregates requests, but batch-invariant kernels eliminate the very optimizations—such as split-K and shape-aware tiling—that make batching effective in the first place. Worse, because kernels are not batch-invariant, adopting this approach requires maintaining a parallel kernel stack solely for determinism. We also quantify the performance cost of this approach below.

\noindent
\textbf{GEMM.} \autoref{fig:motivation:prepej} compares the throughput of \emph{cuBLAS} based GEMMs used in PyTorch against Triton-based batch-invariant kernels developed by He et al. The matrix dimensions correspond to the down projection operation of the \llama model's feed-forward-network. On our GPU, cuBLAS (via \texttt{torch.mm}) reaches up to 527 TFLOPS, whereas the batch-invariant kernel peaks at 194 TFLOPS, a slowdown of 63\%. This gap arises because this Triton-based batch-invariant implementation does not use split-K or exploit newer hardware features such as Tensor Memory Accelerators~\cite{TMA_Engine} or advanced techniques like warp specialization~\cite{fa-3}, all of which are leveraged by PyTorch through vendor-optimized cuBLAS kernels.

\noindent
\textbf{RMSNorm.}
\autoref{fig:motivation:rmsnorm_impl} compares RMSNorm execution time for varying number of tokens for three implementations: Python-based version used in \sglang, Thinking Machines’ Triton-based kernel, and \sglang{}’s default CUDA kernel. The first two are batch-invariant; the CUDA kernel is not. The Python implementation is up to $7\times$ slower than the non-batch-invariant CUDA kernel due to unfused primitive operations and poor shared-memory utilization. The Triton kernel performs substantially better but remains up to 50\% slower than the fused CUDA implementation, which benefits from optimized reductions and kernel fusion. These overheads are amplified at high batch sizes or long context lengths, where normalization may account for a nontrivial fraction of inference time~\cite{gond2025tokenweave}.

\begin{figure}[t!]
    \centering
    \includegraphics[width=0.75\linewidth]{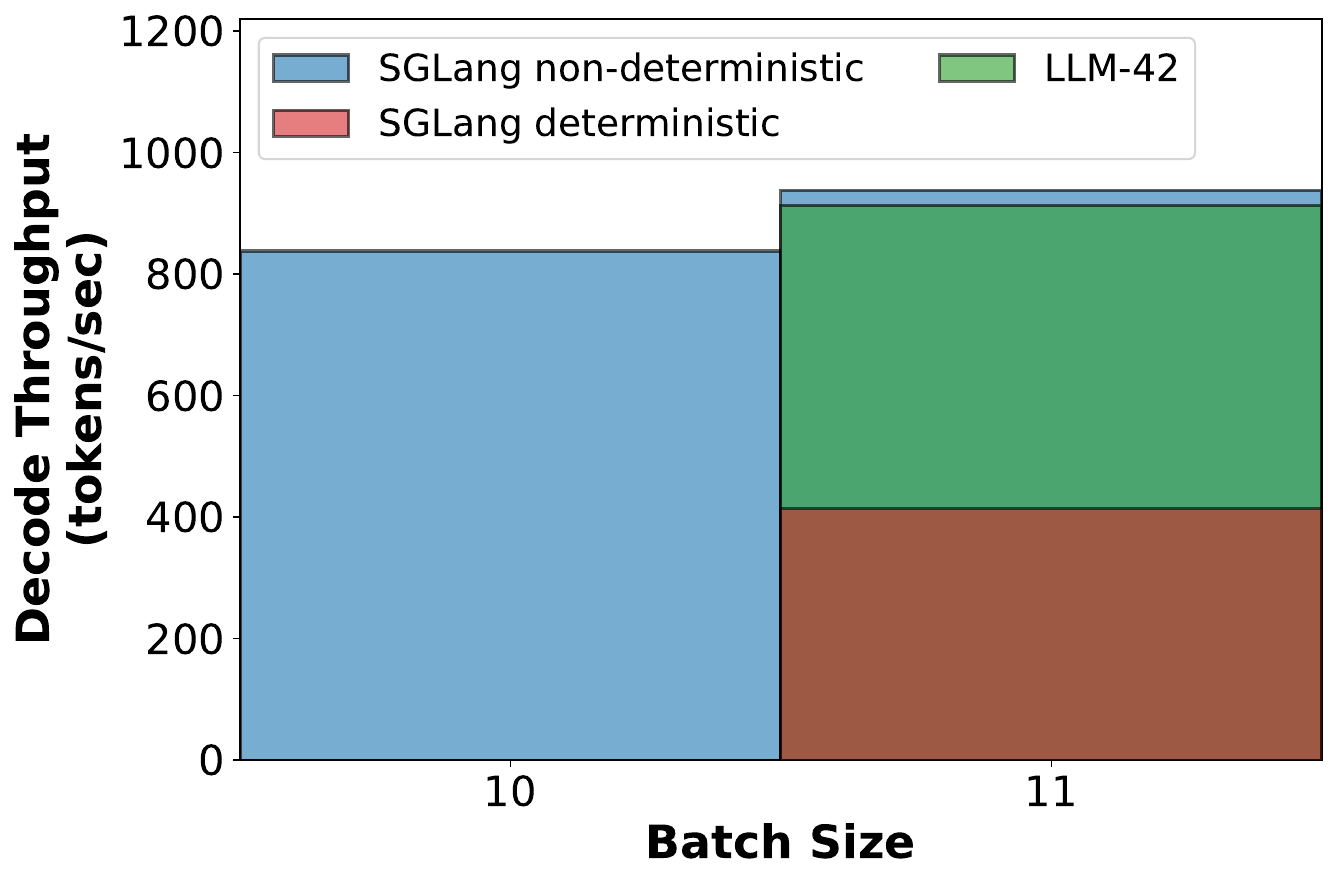}
    \caption{Decode throughput under different scenarios.}
    \label{fig:motivation:decode-tput}
\end{figure}

\noindent
\textbf{End-to-end throughput.}
\autoref{fig:motivation:decode-tput} measures token generation throughput (tokens per second) under three scenarios: (1) 10 requests running in non-deterministic mode, (2) 11 requests running in non-deterministic mode, and (3) 11 requests running in deterministic mode but only one of them requires deterministic output. With 10 concurrent non-deterministic requests, the system generates 845 tokens/s. The batch size increases to 11 when a new request arrives and if decoding continues non-deterministically, throughput improves to 931 tokens/s (a jump of about 10\%). In contrast, if the new request requires determinism, the entire batch is forced to execute through the slower batch-invariant kernels, causing throughput to collapse by 56\% to about 415 tokens/s—penalizing every in-flight request for a single deterministic one. This behavior is undesirable because it couples the performance of all requests to that of the slowest request.

Overall, these results show that batch-invariant execution incurs a substantial performance penalty. While it may be feasible to improve the performance of batch-invariant kernels, doing so would require extensive model- and hardware-specific kernel development. This engineering and maintenance burden makes the approach difficult to sustain in practice. This may be why deterministic inference is largely confined to debugging and verification today, rather than being deployed for real-world LLM serving.

%% file: Sections/3_Observations.tex
\section{Observations}
\label{section:observations}

In this section, we distill a set of concrete observations about non-determinism, GPU kernels and LLM use-cases. These observations expose why batch-invariant computation is overly restrictive and motivate a more general approach to enable determinism in LLM inference.

\begin{tcolorbox}[insightbox]
\textbf{Observation-1 (O1).} \emph{If a sequence is already in a consistent state, the next emitted token is usually consistent even under dynamic batching. However, once a token diverges, autoregressive decoding progressively amplifies this difference over subsequent steps.}
\end{tcolorbox}

This is because tokens become inconsistent only when floating-point drift is large enough to alter the effective decision made by the sampler—e.g., by changing the relative ordering or acceptance of high-probability candidates under the decoder’s sampling policy (e.g., greedy or stochastic sampling). In practice, such boundary-crossing events are rare, as numerical drift typically perturbs logits only slightly. However, autoregressive decoding amplifies even a single such deviation: once a token differs, all subsequent tokens may diverge. Since a single request typically produces hundreds to thousands of output tokens, two sequence-level outputs can look dramatically different even if the initial divergence is caused by a single token flip induced by a different reduction order across runs.

To demonstrate this phenomenon empirically, we conduct an experiment using the \llama model on the \sharegpt dataset. We first execute 350 requests with batch size one—i.e., without dynamic batching—to obtain reference (“ground-truth”) output tokens. We then re-run the same requests under dynamic batching at a load of 6 queries per second and compare each request’s output against its reference. In both runs, we fix the output length to 512 tokens.

\begin{figure}[t!]
    \centering
    \includegraphics[width=0.95\linewidth]{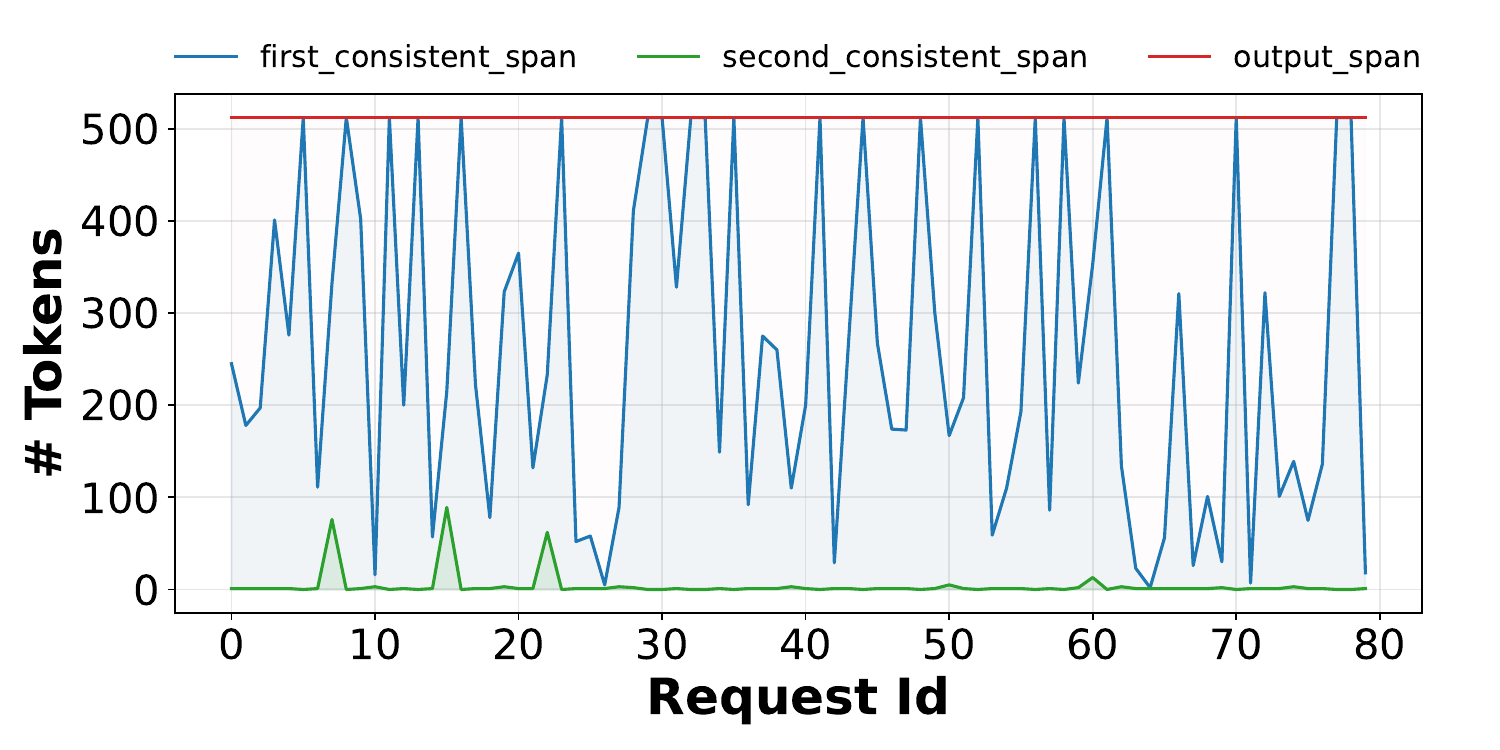}
    \caption{Length of the first and second consistent span (number of tokens that match with the ground-truth) for different requests under dynamic batching.}
    \label{fig:design:drift-stats}
\end{figure}

We quantify divergence using two metrics. The first consistent span of a request measures the number of initial output tokens that match exactly across the two runs, while the second consistent span measures the number of matching tokens between the first and second divergence points. \autoref{fig:design:drift-stats} shows these metrics for 80 requests. In the common case, hundreds of initial tokens are identical across both runs, with some requests exhibiting an exact match of all 512 tokens in the first consistent span. However, once a single token diverges, the sequence rapidly drifts: the second consistent span is near zero for most requests, indicating that divergence quickly propagates through the remainder of the output.

\begin{figure}[t!]
    \centering
    \includegraphics[width=0.95\linewidth]{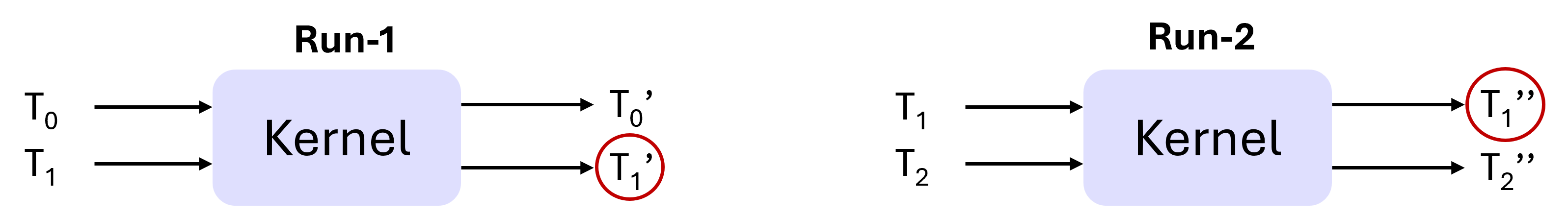}
    \caption{A position-invariant kernel produces the same output for a given input element irrespective of its position in the batch, as long as the total batch size is fixed. In this example, $T_{1}'==T_{1}''$ if the kernel is position-invariant.}
    \label{fig:observation:pi}
\end{figure}

\begin{table}[t]
\centering
\scriptsize
\setlength{\tabcolsep}{4pt}
\begin{tabular}{llcc}
\toprule
\multirow{2}{*}{\textbf{Category}} &
\multirow{2}{*}{\textbf{Operator}} &
\multicolumn{2}{c}{\textbf{Invariant}} \\
\cmidrule(lr){3-4}
& & \textbf{Batch} & \textbf{Position} \\
\midrule

\multirow{1}{*}{Matmul}
& CuBLAS GEMM        & \xmark & \cmark \\
% & Fused MoE (Triton) & \xmark & \cmark \\
\midrule

Attention
& FlashAttention-3$^{\ddagger}$ & \cmark & \cmark \\
\midrule

\multirow{3}{*}{Communication}
& Multimem-based AllReduce$^{*}$ & \cmark & \cmark \\
& Ring-based AllReduce           & \xmark & \xmark \\
& Tree-based AllReduce$^{\star}$ & \cmark & \cmark \\
\midrule

\multirow{2}{*}{Normalization}
& RMSNorm$^{\dagger}$ & \xmark & \cmark \\
& Fused RMSNorm + Residual$^{\dagger}$ & \xmark & \cmark \\
\bottomrule
\end{tabular}
\caption{Invariance properties of common inference operators
($^{\ddagger}$num\_splits=1, $^{*}$CUDA~13.0+, $^{\star}$specific NCCL settings, $^{\dagger}$vLLM/\sglang defaults).}
\label{tab:operator_invariance}
\end{table}

\begin{tcolorbox}[insightbox]
\textbf{Observation-2 (O2).} \emph{Most GPU kernels use uniform, shape-consistent reductions: they apply the same reduction strategy to all elements within a given batch. Moreover, the strategy remains fixed for all batches of the same shape, changing only when the shape changes. }
\end{tcolorbox}

The simplest example of this is GEMM kernels. For a given input matrix A of size M x K, a GEMM kernel computes all the M input elements under the same reduction order (say R). Moreover, it applies the same reduction order R to all input matrices of size M x K. We refer to such kernels as position-invariant. Position invariance implies that, with a fixed total batch size, an input element’s output is independent of its position in the batch.\footnote{Note that such guarantees do not hold for kernels that implement reductions via atomic operations. Fortunately, kernels used in the LLM forward pass do not use atomic reductions.} \autoref{fig:observation:pi} shows an example of a position-invariant kernel and \autoref{tab:operator_invariance} shows the invariance properties of common LLM operators. 

The motivation for batch-invariant computation stems from the fact that GPU kernels used in LLMs, while deterministic for a particular input, are not batch-invariant. We observe that position-invariance captures a strictly stronger property than determinism: determinism only requires that the same input produce the same output across runs, whereas position-invariance implies that the output of a given input remains consistent as long as the input size to the kernel remains the same. This allows us to reason about kernel behavior at the level of input shapes, rather than individual input values.

\begin{tcolorbox}[insightbox]
\textbf{Observation-3 (O3).} \emph{For deterministic inference, it is sufficient to ensure that a given token position goes through the same reduction strategy across all  runs of a given request; reduction strategy for different token positions within or across sequences can be different.}
\end{tcolorbox}

Numerical differences in the output of a token arise from differences in how its own floating-point reductions are performed, not from the numerical values of other co-batched tokens. While batching affects how computations are scheduled and grouped, the computation for a given token position is functionally independent: it consumes the same inputs and executes the same sequence of operations. As a result, interactions across tokens occur only indirectly through execution choices—such as which partial sums are reduced together—not through direct data dependencies. Consequently, as long as a token position  is always reduced using the same strategy, its output remains consistent regardless of how other token positions are computed.

\begin{tcolorbox}[insightbox]
\textbf{Observation-4 (O4).} \emph{Current systems take an all-or-nothing approach: they either enforce determinism for every request or disable it entirely. Such a design is not a natural fit for LLM deployments.}
\end{tcolorbox}

This is because many LLM workloads neither require bit-wise reproducibility nor benefit from it. In fact, controlled stochasticity is often desirable as it  enhances output diversity and creativity of LLMs~\cite{integrating_randomness_llm2024, diversity_bias_llm2025, det-inf-kills}. In contrast, requests such as evaluation runs, safety audits, or regression testing require bit-level reproducibility. Overall, different use-cases  imply that enforcing determinism for all requests is an overkill.

It is also worth highlighting that this all-or-nothing behavior largely stems from the batch-invariant approach that ties determinism to the kernel design. Since all co-batched tokens go through the same set of kernels, determinism becomes a global property of the batch rather than being selective. While one could run different requests through separate deterministic and non-deterministic kernels, doing so would fragment batches, complicate scheduling, and likely hurt efficiency.

\begin{figure*}[t]
    \centering
    \begin{subfigure}{0.48\linewidth}
        \centering
        \includegraphics[trim={0 0 0 5}, clip, width=\linewidth]{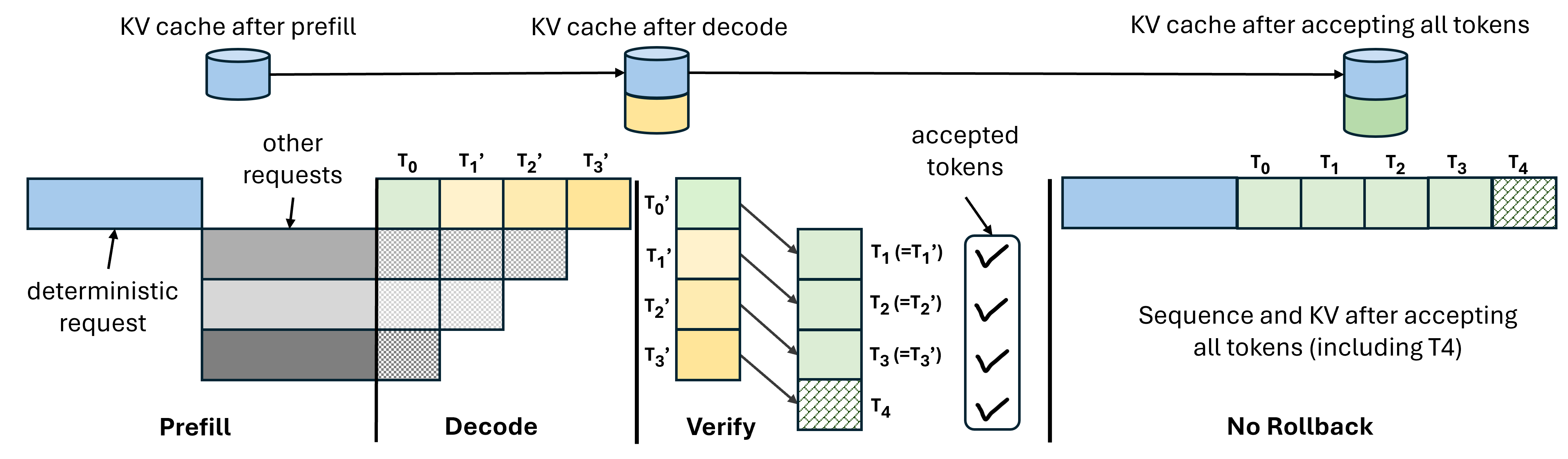}
        \caption{DVR without rollbacks.}
        \label{fig:design:llm42:a}
    \end{subfigure}
    %\\
    \hfill
    \begin{subfigure}{0.48\linewidth}
        \centering
        \includegraphics[trim={0 0 0 0}, clip, width=\linewidth]{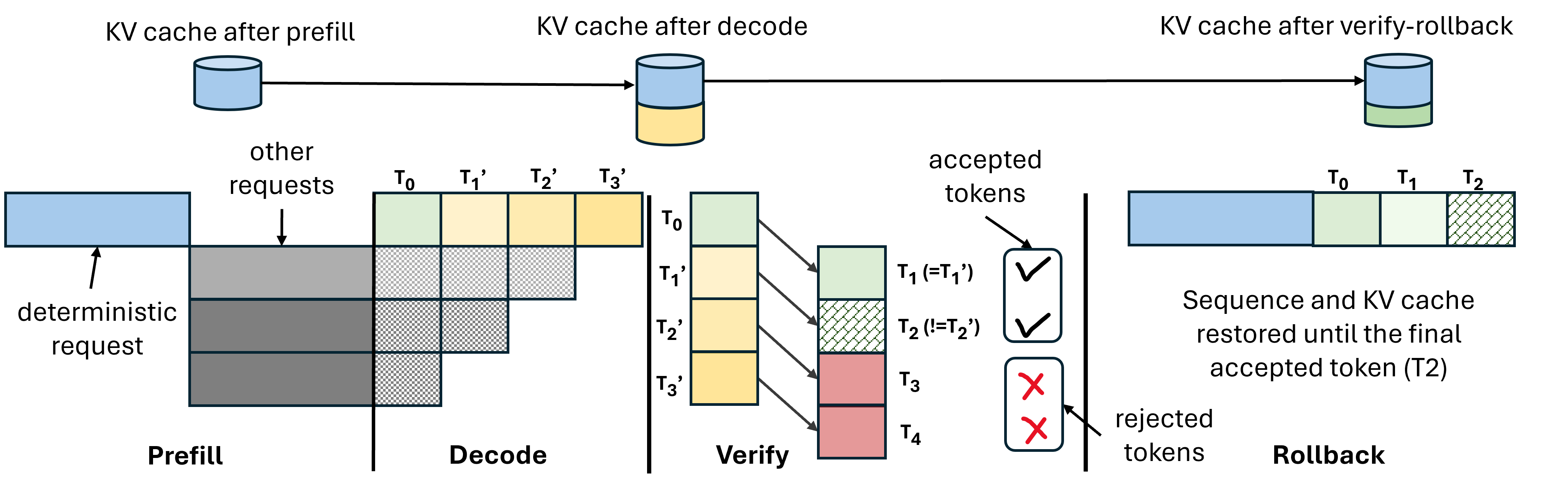}
        \caption{DVR with rollbacks.}
        \label{fig:design:llm42:b}
    \end{subfigure}
    \caption{An example of decode-verify-rollback. After generating a fixed number of tokens through regular decoding, \sysname verifies them in parallel via a separate forward pass. (a) all the tokens generated in the decode phase pass verification; \sysname accepts all these tokens along with the new verifier-generated token ($T_4$). (b) some tokens do not match between decode and verification; \sysname accepts only the initial matching tokens ($T_1'$) and the following verifier-generated token ($T_2$); all other tokens are recomputed. In both cases, the verifier replaces the KV cache entries generated by the decode phase with its own.}
    \label{fig:design:llm42}
\end{figure*}

%% file: Sections/4_Design.tex
\section{\sysname}
\label{sec:design}

Since non-determinism in LLM inference comes from dynamic batching, disabling it would make inference deterministic. However, dynamic batching is arguably the most powerful performance optimization for LLM inference\cite{orca,distserve2024,vllmsosp,sarathi2023}. Therefore, our goal is to enable determinism in the presence of dynamic batching. In this section, we introduce a speculative decoding-inspired, scheduler-driven approach \sysname to achieve this goal.

\subsection{Overall Design}
\sysname exploits the observations presented in~\autoref{section:observations} as follows:

\noindent \textbf{Leveraging O1.} Tokens decoded from a consistent state are mostly consistent. Based on this observation, we re-purpose speculative-decoding style mechanism to enforce determinism via a decode–verify–rollback (DVR) protocol. DVR optimistically decodes tokens using the fast path and a verifier ensures determinism. Only tokens approved by the verifier are returned to the user, while the few that fail verification are discarded and recomputed. The key, however, is to ensure that the verifier’s output itself is always consistent. We leverage O2 to achieve this.

\noindent \textbf{Leveraging O2.} Because GPU kernels rely on shape-consistent reductions, we make the verifier deterministic by always operating on a fixed number of tokens (e.g., verifying 10 tokens at a time). The only corner case occurs at the end of a sequence, where fewer than  T tokens may remain (for instance, when the 11th token is an end-of-sequence token). We handle this by padding with dummy tokens so that the verifier always processes exactly  T tokens.

\noindent \textbf{Leveraging O3.} Determinism only requires that each token position follow a consistent strategy across runs whereas different positions can follow different strategies. This observation lets us compute prefill and decode phases using different strategies. Because prefill is massively parallel even within a single request, it can be made deterministic simply by avoiding arbitrary batching of prefill tokens, eliminating the need for a verifier in this phase. Verifier is required only for the tokens generated by the decode phase.

\noindent \textbf{Leveraging O4:} \sysname decouples determinism from token generation by moving it into a separate verification phase. This makes selective determinism straightforward: only requests that require deterministic inference incur verification, while all other traffic avoids it. We expose this control to the users via a new API flag \texttt{is\_deterministic=True|False} that allows them to explicitly request determinism on a per-request basis; default is \texttt{False}.

\autoref{fig:motivation:decode-tput} quantifies the performance benefit of selective determinism. When one out of 11 requests requires determinism, \sysname achieves decode throughput of 911 tokens per second which is $2.2\times$ higher than the deterministic mode throughput of \sglang and only within 3\% of the non-deterministic mode (best case) throughput. We present a more detailed evaluation in \autoref{sec:evaluation}.

\subsection{Decode-verify-rollback (DVR)}

DVR performs decoding optimistically by first generating tokens using high-throughput, non-deterministic execution and then correcting any inconsistencies through verification and recomputation. Rather than enforcing determinism upfront, it identifies inconsistencies in the generated sequence on the fly, and recomputes only those tokens that are not guaranteed to be consistent across all possible runs of a given request.

\autoref{fig:design:llm42} illustrates how DVR operates, assuming a verification window of four tokens. The blue request requires deterministic output and its first output token $T_0$ is produced by deterministic prefill phase that avoids arbitrary inter-request batching. \sysname then generates candidate tokens $T_1'$, $T_2'$, and $T_3'$ using the regular fast path with dynamic batching, where gray requests may be batched arbitrarily. The first input to the verifier should be consistent (in this case, $T_0$ is consistent since it comes from the prefill phase). The verifier replays these four tokens $T_0$ and $T_1'$–$T_3'$ as input and produces output tokens $T_1$–$T_4$. Verification has two possible outcomes: (1) all tokens pass verification, or (2) one or more tokens fail verification. We describe these two cases in detail below.

\paragraph{Case-1: When verification succeeds.}
In the common case, verification reproduces the same tokens that the preceding decode iterations generated. For example, in~\autoref{fig:design:llm42:a}, $T_1=T_1'$, $T_2=T_2'$ and $T_3=T_3'$. In this case, \sysname accepts all these tokens; in addition, it also accepts the token $T_4$ since $T_4$ was generated by the verifier from a consistent state and is therefore consistent.

\begin{figure*}[t!]
    \centering
        \begin{subfigure}[t]{0.24\linewidth}
        \centering
        \includegraphics[width=\linewidth]{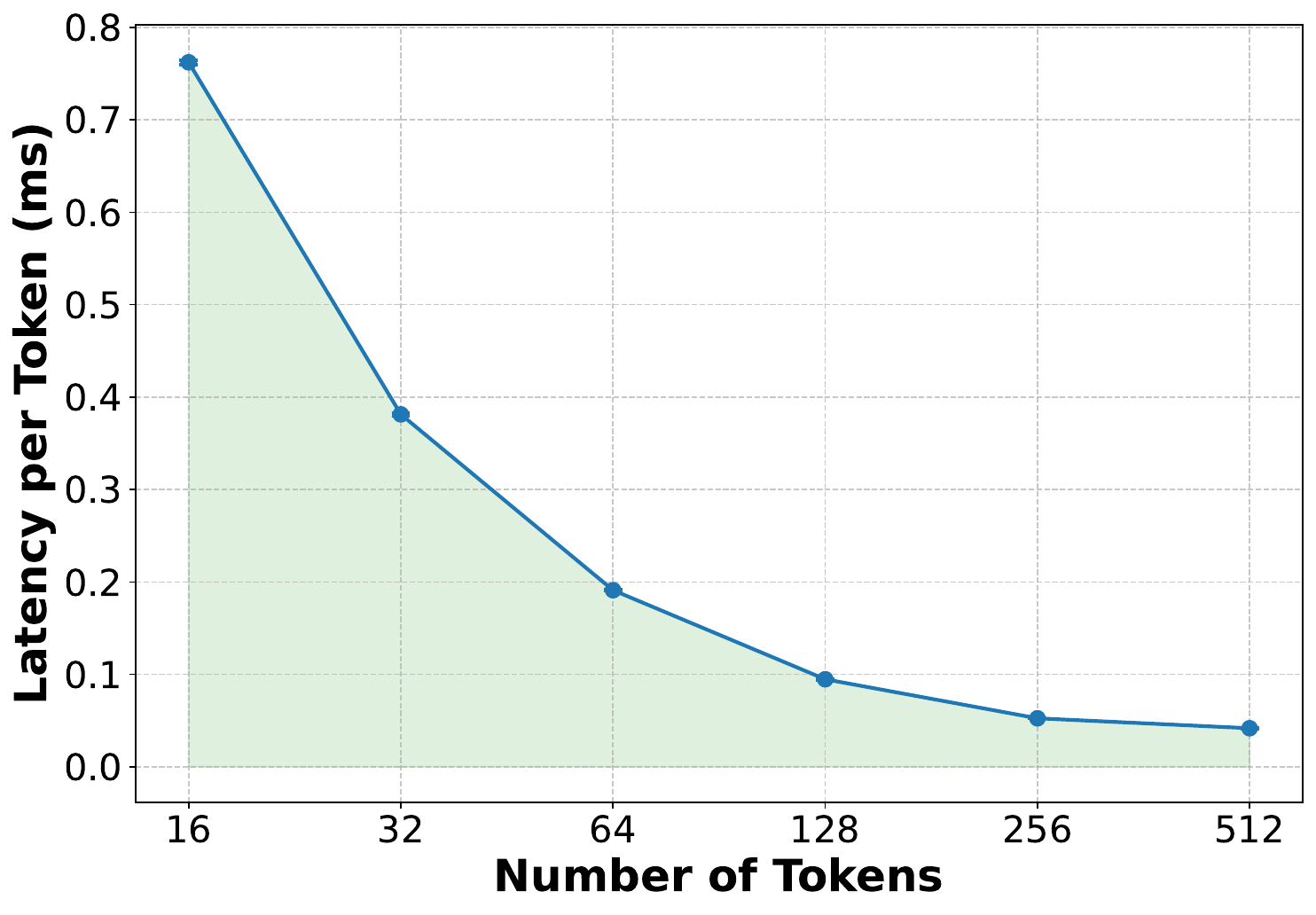}
        \caption{Verification latency}
        \label{fig:design:verification-cost}
    \end{subfigure}\hfill
    \begin{subfigure}[t]{0.24\linewidth}
        \centering
        \includegraphics[width=\linewidth]{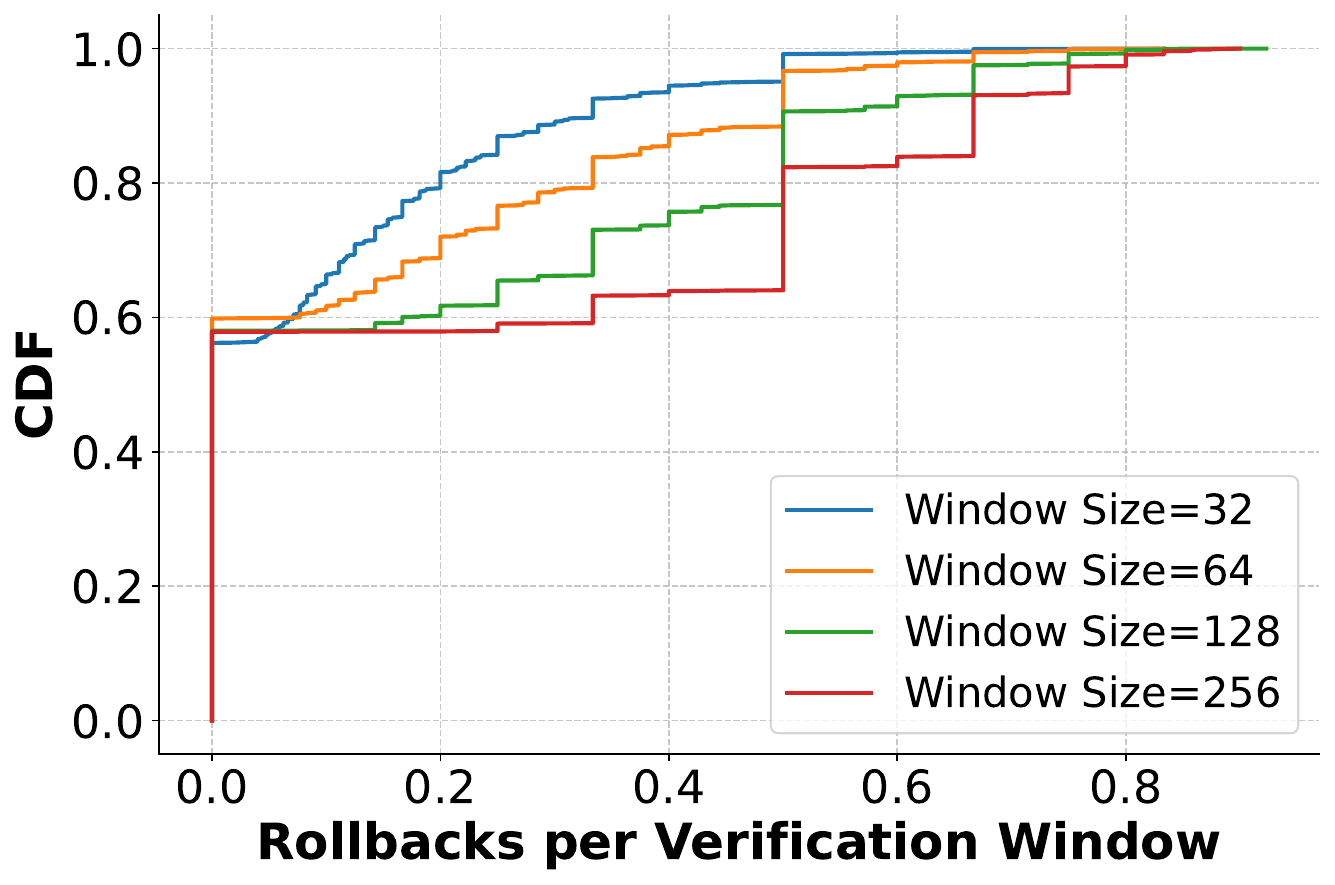}
        \caption{Rollbacks}
        \label{fig:design:rollbacks}
    \end{subfigure}\hfill
    \begin{subfigure}[t]{0.24\linewidth}
        \centering
        \includegraphics[width=\linewidth]{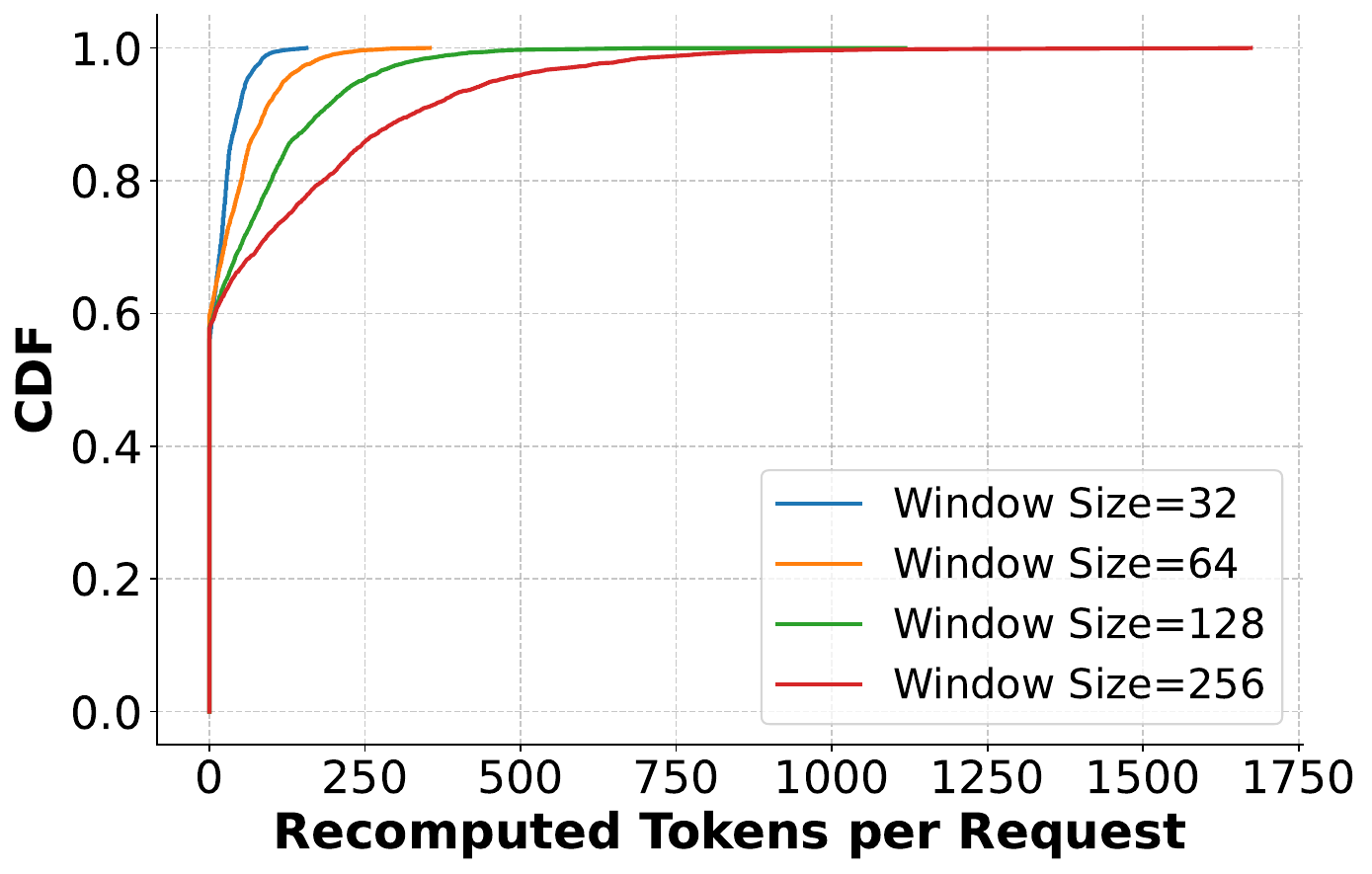}
        \caption{CDF of recomputed tokens}
        \label{fig:design:recomputation:cdf}
    \end{subfigure}\hfill
    \begin{subfigure}[t]{0.24\linewidth}
        \centering
        \includegraphics[width=\linewidth,height=2.9cm]
        {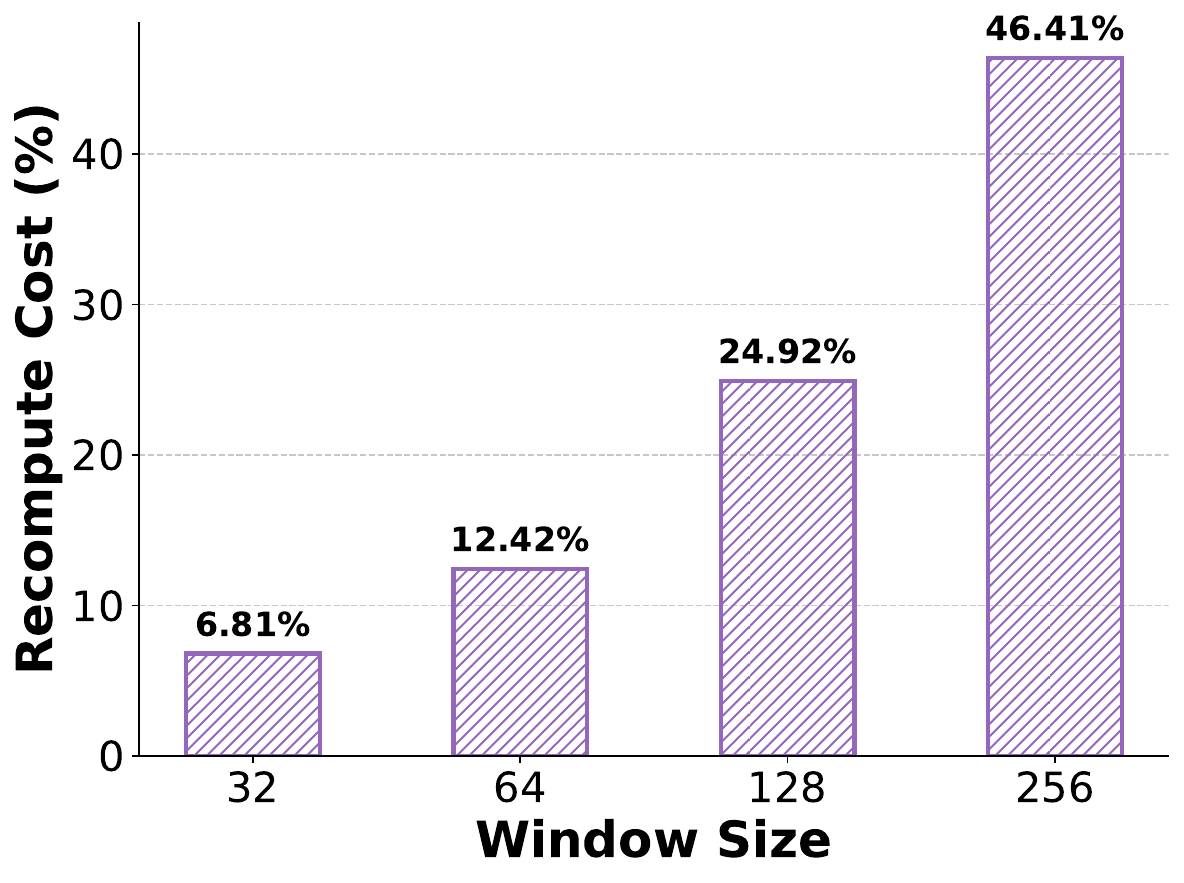}
        \caption{Total recomputation overhead}
        \label{fig:design:recomputation:total}
    \end{subfigure}
    \caption{The cost of verification and recomputation with varying window sizes.}
    \label{fig:design:grouped-verification:online}
\end{figure*}

\paragraph{Case-2: When verification fails.}
Occasionally, a verification pass disagrees with one or more decoded tokens, e.g., only $T_1'$ matches with the verifier's output in~\autoref{fig:design:llm42:b}. In this case, \sysname commits the output only up to the last matching token ($T_1'$); all subsequent tokens ($T_2'$ and $T_3'$) are rejected. The verifier-generated token that appears immediately after the last matching position ($T_2$) is also accepted and the KV cache is truncated at this position. The next decode iteration resumes from this repaired, consistent state.  

\paragraph{Making KV cache consistent.} During the decode phase, the KV cache is populated by fast-path iterations that execute under dynamic batching. Consequently, even when token verification succeeds, the KV cache corresponding to the verified window may still be inconsistent, since it was produced by non-deterministic execution. This inconsistency could affect tokens generated in future iterations despite the current window being verified. Token-level verification alone is therefore insufficient without repairing the KV cache. To address this, we overwrite the KV cache entries produced during decoding with the corresponding entries from the verification pass. This ensures that both the emitted tokens and the KV cache state are consistent for subsequent decode iterations, eliminating downstream divergence.

\paragraph{Guaranteed forward progress.} Note that each verification pass produces at least one new consistent output token as shown in~\autoref{fig:design:llm42}. As a result, DVR guarantees forward progress even in contrived worst-case scenarios that might trigger rollbacks for every optimistically generated token. We have not observed any such scenario in our experiments.

Overall, DVR follows the high-level structure of speculative decoding, but differs in several important ways as discussed in \autoref{tab:speculative_vs_llm42}. Under DVR, decoded tokens result from a faithful execution of the model’s forward pass, with deviations arising only from the floating-point rounding errors. In contrast, speculative execution techniques generate candidate tokens using modified forward computations—for example, by running a smaller or distilled draft model~\cite{specdecoding-icml2023,specinfer-2024} or by truncating or pruning attention/context~\cite{medusa,fu2023lookahead}. Because these approximations change the computation itself, speculation depth is typically limited to a few (2-8) tokens~\cite{medusa,fu2023lookahead,specdecoding-icml2023,specinfer-2024}. In contrast, leveraging O1, DVR can safely speculate over longer (32-64) token sequences. For the same reason, verification in DVR succeeds with high probability, in contrast to the much lower acceptance rates observed in existing speculative decoding schemes. Finally, DVR performs verification using the same model, whereas most speculative decoding approaches require a separate model for verification.

\subsection{Grouped Verification}
\label{sec:design:grouped-verification}

The efficiency of DVR depends on the size of verification window and involves a trade-off: smaller windows incur higher verification cost but low recomputation overhead, whereas larger windows reduce verification cost at the expense of increased recomputation. We empirically characterize this trade-off by profiling \llama on an H100-PCIe GPU. We measure per-token verification cost based on the latency of verification forward pass. To quantify the recomputation cost, we execute 4096 requests from the ShareGPT dataset in an online setting at 12 queries per second. For each window size, we measure the total number of tokens decoded versus the number of tokens returned to the user; the difference between the two denotes recomputation overhead.

\autoref{fig:design:verification-cost} shows the per-token verification cost as a function of verification window size. 
For smaller windows, the verification pass is memory-bound: each token performs very little computation, resulting in low arithmetic intensity and poor hardware utilization. Consequently, the verification cost is high—up to 0.75 ms per-token. As the verification window increases, the kernel transitions to being compute-bound, and the per-token cost drops sharply, reaching 0.05 ms at a window size of 512, signifying a reduction of $15\times$.

\autoref{fig:design:rollbacks} shows that more than half of the requests complete without any rollback (i.e., all their output tokens pass verification), while a small fraction of requests incur frequent rollbacks. Moreover, rollbacks become more common as the window size increases. For example, with a window size of 256, about 40\% of requests observe a rollback ratio of 50\% or higher indicating that at least half of the verification steps detect one or more mismatched tokens. In contrast, with a window size of 32, only about 5\% of requests reach this level. The number of recomputed tokens per request also follows the same trend as shown in \autoref{fig:design:recomputation:cdf}. Further, \autoref{fig:design:recomputation:total} shows the resulting total recomputation overhead. Note that a verification failure requires recomputing all tokens following the mismatched position within the current window. As a result, larger windows incur higher recomputation overhead on average. Concretely, recomputation overhead is 6.81\% at a window size of 32 and increases roughly linearly, reaching 46.41\% at a window size of 256.

We propose \textit{grouped verification} to address the inherent trade-off between the cost of verification and recomputation. Instead of verifying a large window of a single request (256 tokens), we verify small, fixed-size windows of multiple requests together in a single pass (e.g., 8 requests, 32 tokens each). This way each request retains the rollback properties of a small window, while the verification pass operates on a large effective batch, achieving high utilization and low verification cost.

\subsection{Discussion}
\label{sec:design:op-discussion}

Guaranteeing determinism requires every operator in the inference pipeline to behave consistently. This section highlights a few subtle sources of inconsistency in commonly used operators. We do not introduce new techniques here; instead, we adopt established approaches to enforce consistent behavior and summarize them for completeness.

\paragraph{Attention.}
Attention involves reductions along multiple steps: reductions along the sequence dimension (e.g., in FlashDecoding-style sequence parallelism~\cite{flashdecoding}) and reductions inside the softmax computation. To make it consistent, we use FlashAttention-3 attention kernel and set the number of KV splits to one in the verification pass; fast-path decode iterations run as usual.\footnote{Some performance optimizations are possible based on \sglang implementation \url{https://lmsys.org/blog/2025-09-22-sglang-deterministic/}.}

\paragraph{Communication collectives.}
Classical ring-based AllReduce reduces tokens in different orders depending on their position in the batch. In contrast, multimem-based AllReduce, introduced in CUDA 13.0, follows consistent reduction schedules~\cite{nvls-deterministic} and should be preferred whenever supported. On older platforms where multimem/NVLS is unavailable, one can use tree-based AllReduce with fixed NCCL configuration (e.g., setting num\_channels to one and using a fixed protocol) to achieve determinism, at the expense of higher communication cost than ring-based AllReduce.

\paragraph{Sampling.}
We adopt sampler from \sglang. When temperature is zero, it selects the token with the highest logit (argmax) and if there are multiple maximal values then it returns the index of the first maximal value. For non-greedy sampling, however, the sampling process involves randomness, making the outputs sensitive to how random numbers are generated and consumed. To address this issue, \sglang introduces a new sampling function, \verb!multinomial_with_seed!, which replaces \verb!torch.multinomial!, an inherently non-deterministic operator under batched execution. This function perturbs logits with Gumbel noise generated from a seeded hash function, ensuring that the same input–seed pair always produces the same sample\cite{SGLangTeam2025}.

Overall, enforcing determinism across the entire inference pipeline requires careful configuration of the attention kernel and selecting an appropriate AllReduce implementation. The sampling module does require a new implementation, but this is a one-time cost since the same sampler is used across all models. In contrast, the bulk of the performance and engineering effort in LLM systems lies in GEMM kernels, which \sysname reuses entirely, along with RMSNorm and FusedMoE kernels.

%% file: Sections/5_Evaluation.tex
\section{Evaluation}
\label{sec:evaluation}

Our evaluation answers the following questions:

\begin{table}[t]
\centering
\small
\resizebox{\linewidth}{!}{%
\begin{tabular}{lllccc}
\toprule
\textbf{Dataset} & \textbf{\# Requests} & \textbf{Type} & \textbf{Mean} & \textbf{Median} & \textbf{Std. Dev.} \\
\midrule
\textbf{ShareGPT} & 92812 & Input Length  & 304 & 136 & 491 \\
                  &       & Output Length & 192   & 118   & 212   \\
\textbf{ArXiv}    & 5941  & Input Length  & 7017 & 6435 & 3479 \\
\textbf{Test Split}    &       & Output Length & 198   & 191   & 74   \\
\bottomrule
\end{tabular}%
}
\caption{Datasets and their input/output context lengths.}
\label{tab:dataset_stats}
\end{table}

\begin{figure*}[t!]
    \centering
    \includegraphics[width=0.98\linewidth]{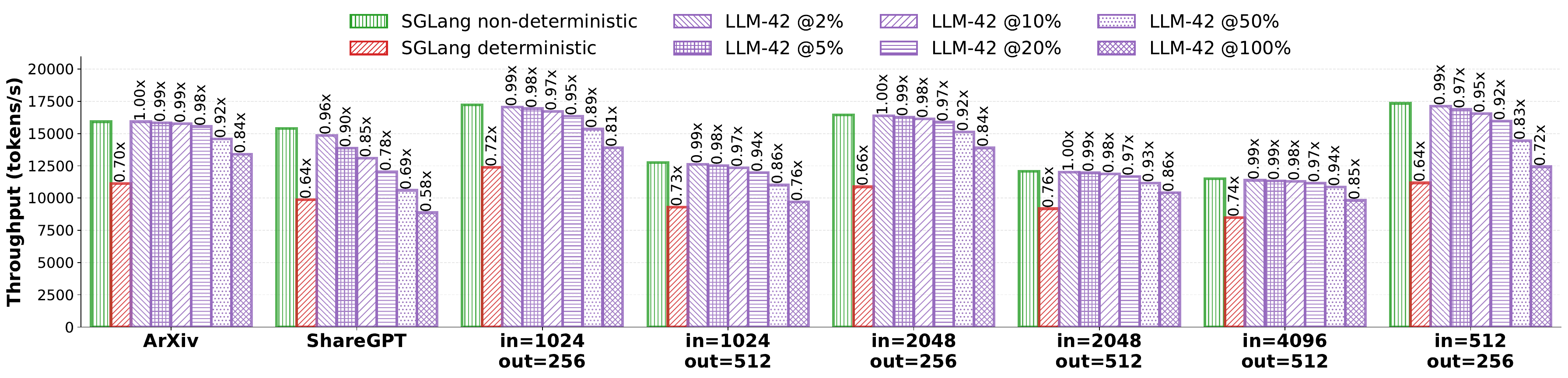}
    \caption{Throughput in offline inference. \sglang has only two modes: all requests run in either deterministic or non-deterministic mode whereas \sysname supports selective determinism (\% values in the legend reflect the fraction of traffic that requires deterministic output).}
    \label{fig:eval_offline}
\end{figure*}

\begin{itemize}
    \item How does \sysname compare against the baseline's deterministic and non-deterministic modes?
    \item How does \sysname perform for different mix of deterministic and non-deterministic traffic?
    \item How do configuration parameters affect the performance of grouped verification in \sysname?
\end{itemize}

\running{Models and environment} We evaluate \sysname with the \llama model. It has 32 query heads, 8 KV heads, and 32 layers. Experiments were conducted on a system equipped with four NVIDIA H100 PCIe GPUs with 80 GB HBM3 memory and 114 streaming multiprocessors per GPU. The host CPU has 64 physical cores (128 hardware threads) and approximately 1.65 TB of DRAM.

\running{Workloads} 
We evaluate on synthetic inputs with varying context lengths, following common practice in prior work~\cite{nanoflow2024}. We additionally benchmark on the widely used ShareGPT~\cite{huggingfacesharegpt} and Arxiv~\cite{arxiv-summarization} datasets, whose characteristics are summarized in Table~\ref{tab:dataset_stats}. All experiments use the \verb!meta-llama/Llama-3.1-8B-Instruct! tokenizer. To evaluate \sysname under different workload conditions, we vary the fraction of requests that require deterministic output between 2\%, 5\%, 10\%, 20\%, 50\%, and 100\%. While higher deterministic ratios are included for stress testing, we expect that in practical deployments only a small fraction of requests will require determinism. Finally, note that the deterministic request ratio applies only to \sysname; approaches based on batch-invariant computation support only global determinism.

\running{Serving system baselines} We implement \sysname on top of the \sglang serving framework version 0.5.3rc0 and compare performance against \sglang with deterministic execution switched on. To understand the upper limit on performance, we also compare with the non-deterministic version of \sglang. We refer to these two baselines as \sglangdet and \sglangnondet. We use FA-3 attention kernel in all our experiments; we set \verb!num_splits = 1! in the verification step of LLM-42 and use the default settings for the decode phases.

\running{Metrics} In offline inference, we evaluate throughput as the number of tokens processed per second. For online inference, we evaluate end-to-end request execution latencies and time-to-first-token. To better understand the overhead of our approach, we also report the number of rollbacks and recomputed tokens across different configurations of \sysname.

\subsection{Offline Inference}

For offline inference, we evaluate eight workload configurations: two traces from the ArXiv and ShareGPT datasets and six other configurations with different input and output lengths. Each configuration executes 4096 requests to completion. \autoref{fig:eval_offline} reports the resulting throughput across systems measured in tokens per second; we summarize the main observations below.

Enabling deterministic inference in \sglang incurs a substantial throughput penalty, ranging from 24\% (in=2048, out=512) to 36\% (ShareGPT). This slowdown stems from the use of batch-invariant kernels, which are consistently slower than the standard optimized kernels, as shown in \autoref{fig:motivation:operator-perf-combined}. In contrast, \sysname mostly uses regular optimized kernels and therefore achieves significantly higher throughput. Even in its worst case—when 100\% of requests require deterministic output—\sysname outperforms \sglangdet in all but one setting (ShareGPT), where it is only 6\% slower.

\begin{table}[!t]
\centering
\scalebox{0.75}{
\begin{tabular}{lcccccc}
\toprule
& \multicolumn{6}{c}{\textbf{Deterministic Ratio}} \\
\cmidrule(lr){2-7}
\textbf{Dataset / Config} & 2\% & 5\% & 10\% & 20\% & 50\% & 100\% \\
\midrule
\multicolumn{7}{c}{\textit{Total number of rollbacks}} \\
\midrule
ArXiv & 70 & 170 & 372 & 697 & 1733 & 3351 \\
ShareGPT & 4 & 5 & 10 & 15 & 44 & 96 \\
in=512, out=256   & 0 & 0 & 0 & 0 & 0 & 0 \\
in=1024, out=256  & 0 & 1 & 5 & 5 & 9 & 24 \\
in=1024, out=512  & 48 & 106 & 225 & 386 & 792 & 1536 \\
in=2048, out=256  & 7 & 22 & 79 & 134 & 378 & 724 \\
in=2048, out=512  & 0 & 0 & 0 & 2 & 2 & 4 \\
in=4096, out=512  & 31 & 79 & 98 & 220 & 538 & 1087 \\
\midrule
\multicolumn{7}{c}{\textit{Total number of recomputed tokens}} \\
\midrule
ArXiv & 1805 & 4414 & 8737 & 13198 & 37687 & 89248 (10.97\%) \\
ShareGPT & 139 & 158 & 164 & 396 & 1185 & 2691 (0.32\%) \\
in=512, out=256   & 0 & 0 & 0 & 0 & 0 & 0 (0\%) \\
in=1024, out=256  & 0 & 44 & 151 & 151 & 288 & 776 (0.07\%) \\
in=1024, out=512  & 1724 & 3272 & 6615 & 12392 & 25967 & 50454 (2.41\%) \\
in=2048, out=256  & 268 & 767 & 2470 & 4080 & 12173 & 21772 (2.08\%) \\
in=2048, out=512  & 0 & 0 & 0 & 81 & 81 & 101 ($<\!0.01\%$) \\
in=4096, out=512  & 1134 & 2961 & 3598 & 7208 & 17161 & 32430 (1.55\%) \\
\bottomrule
\end{tabular}
}
\caption{Rollback and recomputation statistics (grouped verification over 8 requests, 64 tokens each). Recompute cost is shown only for the column with 100\% deterministic traffic.}
\label{tab:eval:rollback_recompute_ws_64_bs_8}
\end{table}

Crucially, \sysname’s throughput improves monotonically as the fraction of deterministic requests decreases. On ArXiv, \sysname operates within 8\%, 2\%, 1\% and 1\% of the best-case throughput (\sglangnondet) at 5\%, 10\%, 20\%, and 50\% deterministic traffic, respectively. This behavior is expected: \sysname imposes no computation overhead on requests that do not require determinism. As a result, \sysname is up to 41\% faster than \sglangdet in these regimes. This trend holds across all workload configurations; for example, at 10\% deterministic traffic, \sysname is 33\% faster than \sglangdet on ShareGPT and up to 48\% faster (in=2048, out=256) on other workloads.

\autoref{tab:eval:rollback_recompute_ws_64_bs_8} reports two complementary metrics that characterize the overhead of enforcing determinism in \sysname: (1) the total number of rollbacks aggregated over all 4096 requests, and (2) the total number of recomputed tokens incurred due to these rollbacks. We find that some configurations incur zero rollbacks even under non-trivial deterministic ratios, and the average case remains low across datasets and input–output lengths. Even in the worst case, ArXiv at 100\% deterministic traffic, the system triggers 3351 rollbacks over 4096 requests, i.e., fewer than one rollback per request on average. A similar trend holds for recomputation overhead: the recomputed token fraction is consistently small, with the worst case at 100\% deterministic traffic reaching at most 10.97\%, while the average case across datasets and configurations is substantially lower. Overall, these results indicate that both rollback frequency and recomputation cost are modest in practice.

\begin{figure*}[t!]
    \centering
    \begin{subfigure}[t]{0.24\linewidth}
        \centering
        \includegraphics[width=\linewidth]{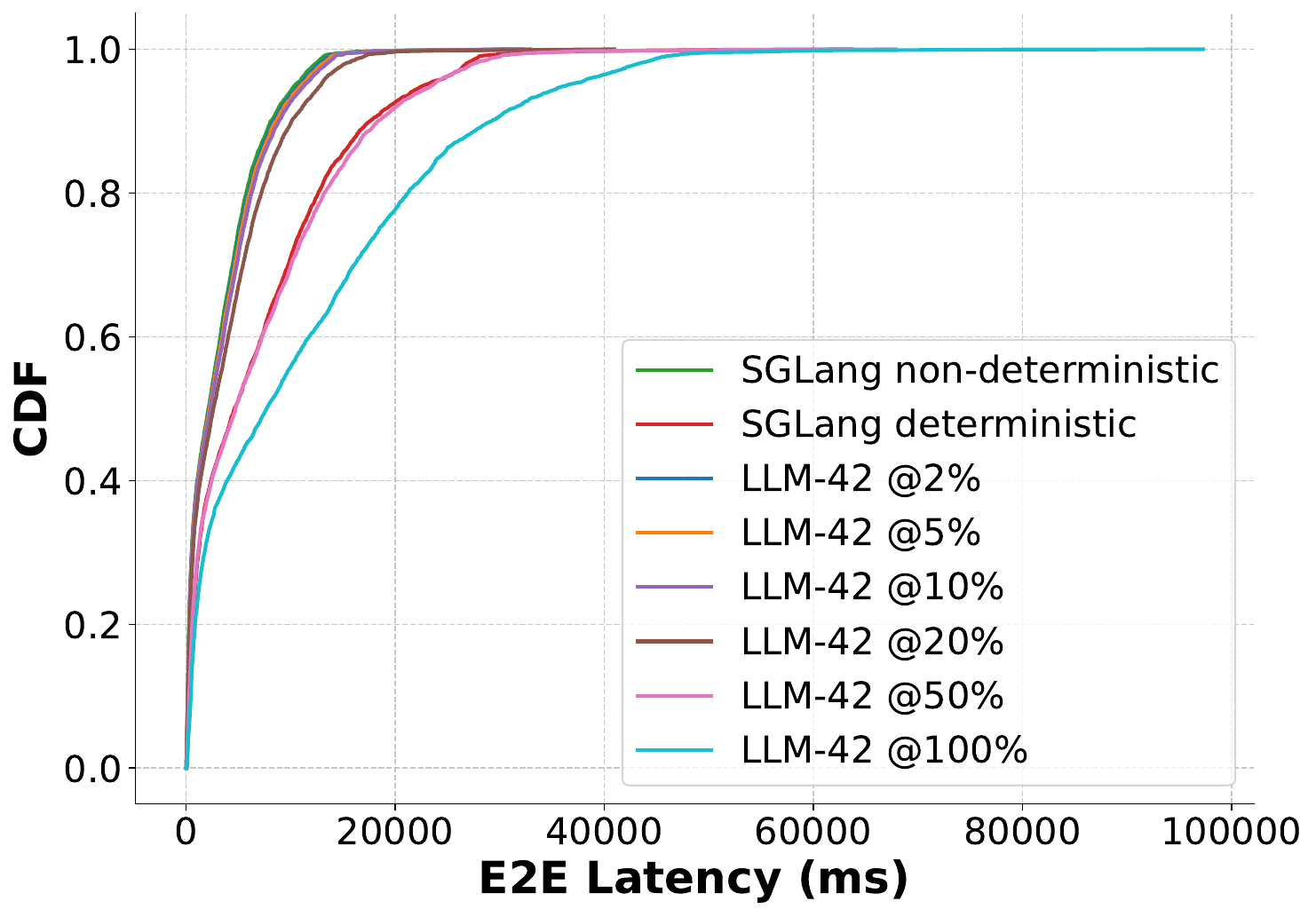}
        \caption{QPS=12}
    \end{subfigure}\hfill
    \begin{subfigure}[t]{0.24\linewidth}
        \centering
        \includegraphics[width=\linewidth]{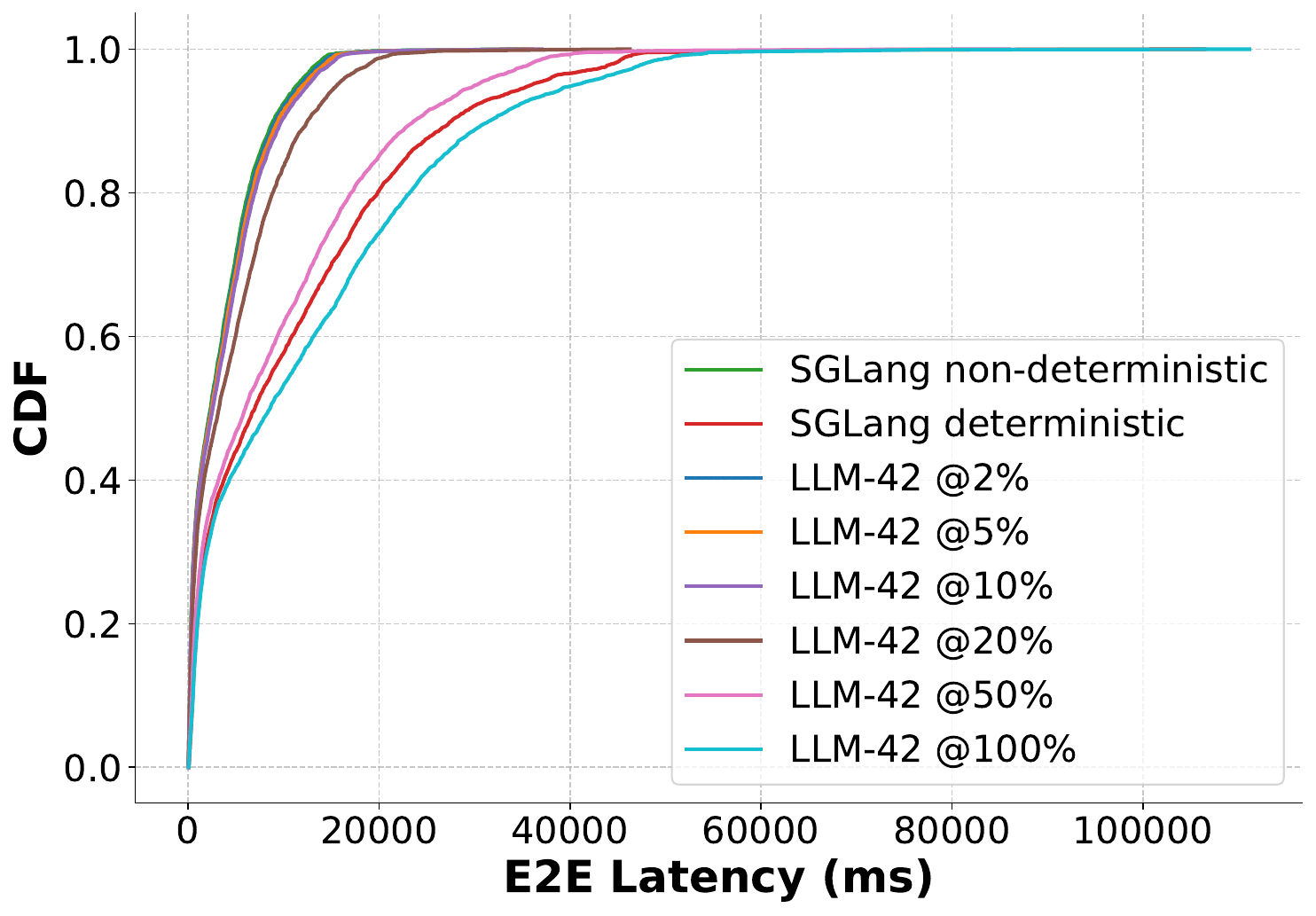}
        \caption{QPS=14}
    \end{subfigure}\hfill
    \begin{subfigure}[t]{0.24\linewidth}
        \centering
        \includegraphics[width=\linewidth]{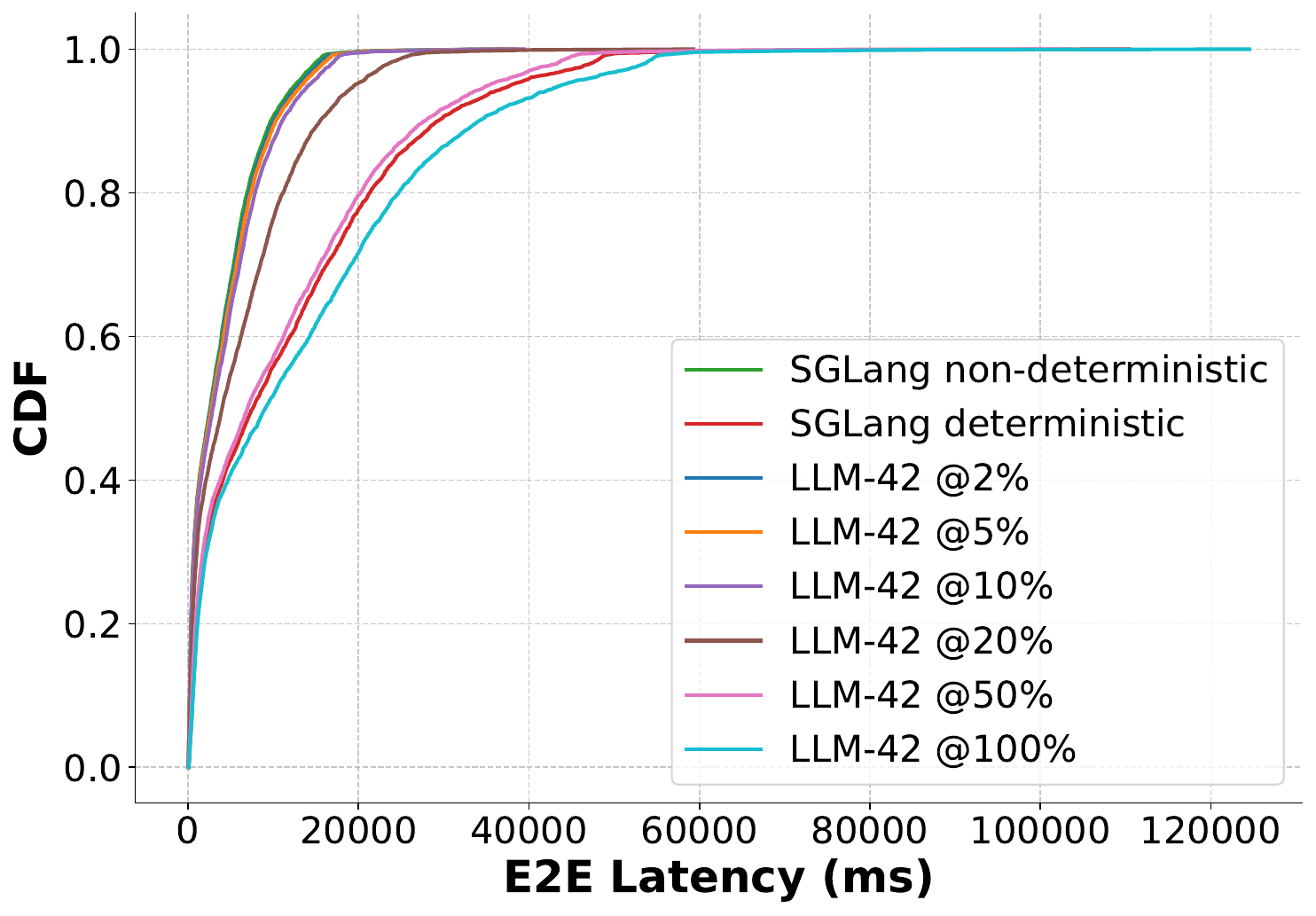}
        \caption{QPS=16}
    \end{subfigure}\hfill
    \begin{subfigure}[t]{0.24\linewidth}
        \centering
        \includegraphics[width=\linewidth]{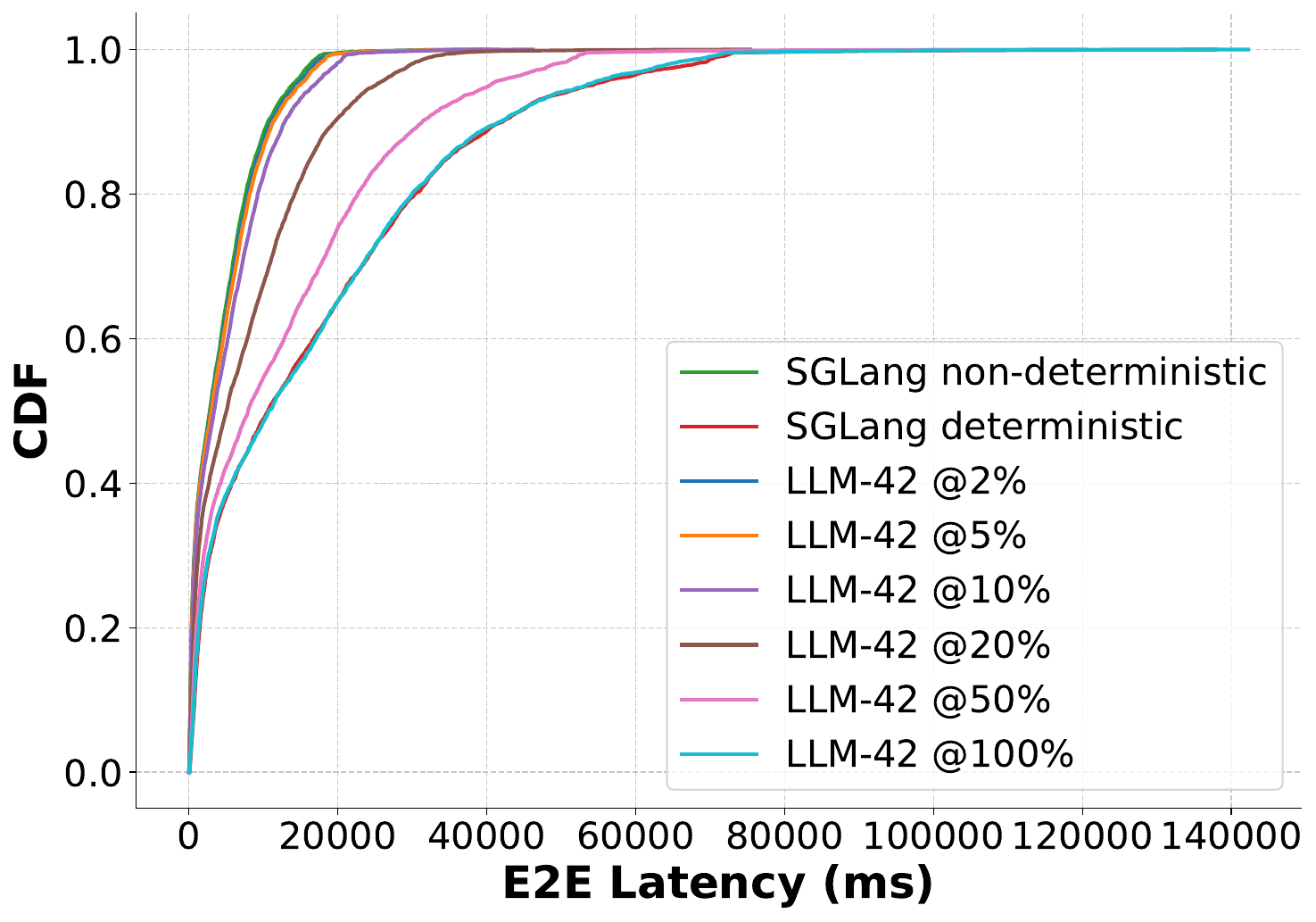}
        \caption{QPS=18}
    \end{subfigure}
    \caption{CDF of request end-to-end latency in online inference with varying load for the ShareGPT dataset.}
    \label{fig:eval:online:sharegpt:e2e}
\end{figure*}

\subsection{Online Inference}
\begin{table*}[t]
\centering
\resizebox{0.9\textwidth}{!}{%
\begin{tabular}{llcccccccc}
\toprule
\textbf{QPS} & \textbf{TTFT} & \textbf{\sglang-Non-Det} & \textbf{\sglang-Det} & \textbf{\sysname@2\%} & \textbf{\sysname@5\%} & \textbf{\sysname@10\%} & \textbf{\sysname@20\%} & \textbf{\sysname@50\%} & \textbf{\sysname@100\%} \\
\midrule
\midrule
\multirow{3}{*}{12} & P50 & 27.0 & 53.4 & 27.4 & 27.6 & 27.9 & 29.3 & 38.5 & 45.4 \\
 & P75 & 35.4 & 69.8 & 35.8 & 36.1 & 37.1 & 41.1 & 52.0 & 58.0 \\
 & P90 & 50.3 & 107.8 & 52.5 & 54.1 & 54.0 & 57.8 & 68.1 & 73.9 \\
\midrule
\multirow{3}{*}{14} & P50 & 27.9 & 60.1 & 27.9 & 28.2 & 28.9 & 31.1 & 42.5 & 48.2 \\
 & P75 & 37.0 & 79.9 & 37.5 & 37.8 & 39.5 & 45.1 & 57.0 & 61.6 \\
 & P90 & 55.8 & 127.0 & 57.0 & 56.5 & 58.6 & 63.8 & 75.8 & 82.0 \\
\midrule
\multirow{3}{*}{16} & P50 & 28.8 & 62.5 & 28.8 & 29.2 & 29.8 & 35.2 & 46.1 & 51.2 \\
 & P75 & 39.0 & 86.0 & 39.7 & 40.5 & 41.6 & 51.2 & 61.0 & 65.9 \\
 & P90 & 60.5 & 138.6 & 61.7 & 60.9 & 63.3 & 74.0 & 83.6 & 87.8 \\
\midrule
\multirow{3}{*}{18} & P50 & 29.8 & 76.2 & 30.2 & 30.6 & 32.2 & 40.7 & 50.7 & 57.1 \\
 & P75 & 41.4 & 105.6 & 42.7 & 43.3 & 46.3 & 57.5 & 67.3 & 75.4 \\
 & P90 & 65.2 & 171.6 & 65.9 & 67.6 & 70.2 & 79.3 & 90.5 & 101.2 \\
\bottomrule
\end{tabular}%
}
\caption{Time-to-first-token (TTFT) latency with varying load for the ShareGPT dataset.}
\label{tab:ttft_latency}
\end{table*}

\autoref{fig:eval:online:sharegpt:e2e} reports the CDF of end-to-end latency for online inference under increasing load on the ShareGPT dataset, with each experiment running 4096 requests. Across all QPS (queries per second) values, \sglangdet exhibits a pronounced rightward shift with a long tail, reflecting substantially higher median and tail latencies. For example, at 12 QPS, \sglangdet{}’s median (P50) latency is 4.64 seconds with a P99 of 28 seconds, compared to 2.15 seconds (P50) and 13.2 seconds (P99) for \sglangnondet. This gap widens further under higher load: at 18 QPS, \sglangdet reaches a P50 latency of 10.6 seconds and a P99 latency of 71.1 seconds, whereas \sglangnondet remains at 2.84 and 17.4 seconds, respectively. \sglangnondet consistently achieves the lowest latency and thus serves as a practical lower bound. In contrast, \sysname closely tracks the non-deterministic baseline, with only modest increases in latency as the fraction of deterministic traffic rises from 2\% to 100\%. At 12 QPS, \sysname at 2\% deterministic traffic incurs only a 3\% increase in median latency over \sglangnondet (2.21 vs. 2.15 seconds), and even at 50\% deterministic traffic, its P99 latency is comparable to that of \sglangdet (at low load) or better (at higher load). This degradation is smooth and monotonic across all loads: at the highest QPS, \sysname maintains significantly tighter CDFs and substantially lower tail latency than \sglangdet. Only at lower QPS and when most requests require deterministic output (100\%) does \sysname exhibit higher latency than the deterministic baseline. This behavior stems from two implementation artifacts: (1) verification currently induces a global pause that temporarily stalls all in-flight requests, and (2) prefill is not batched in our current prototype, reducing efficiency for short input prompts. We plan to address them as part of future work.

Across all QPS levels, time-to-first-token (TTFT) latency in \sysname also increases monotonically with the fraction of deterministic traffic, with modest overhead at low ratios (2–10\%) and higher once deterministic traffic exceeds ~20–50\%. However, even when the entire traffic is deterministic, \sysname still provides much lower tail TTFT than \sglangdet: at QPS 18, P90 TTFT of \sysname is 101.2 milliseconds vs. 171.6 milliseconds of \sglangdet.

\subsection{Ablation Study}
\begin{figure}[t!]
    \centering
    \begin{subfigure}[t]{0.49\columnwidth}
        \centering
        \includegraphics[width=\linewidth]{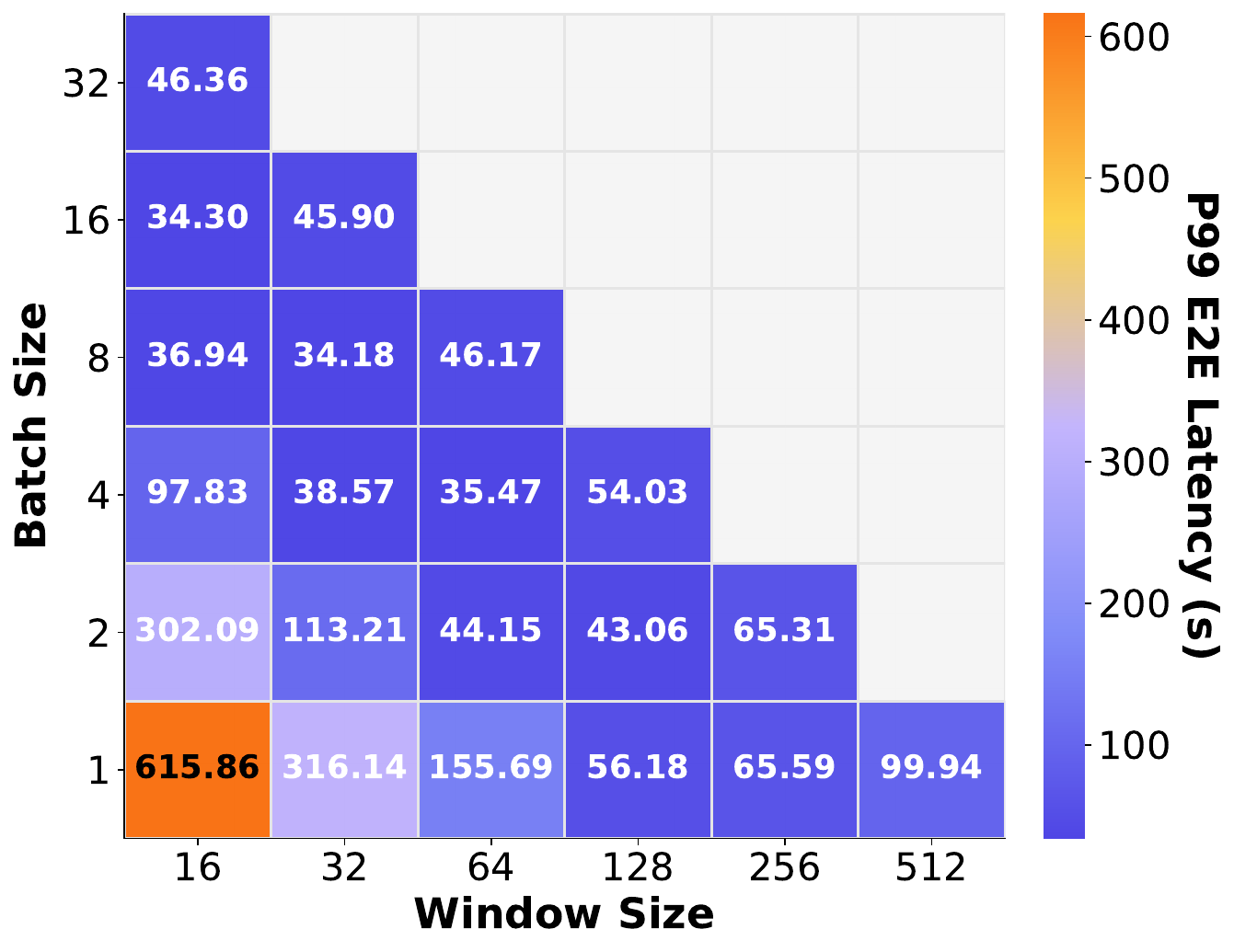}
        \caption{P99 latency}
        \label{fig:eval:ablation:latency}
    \end{subfigure}
    \hfill
    \begin{subfigure}[t]{0.49\columnwidth}
        \centering
        \includegraphics[width=\linewidth]{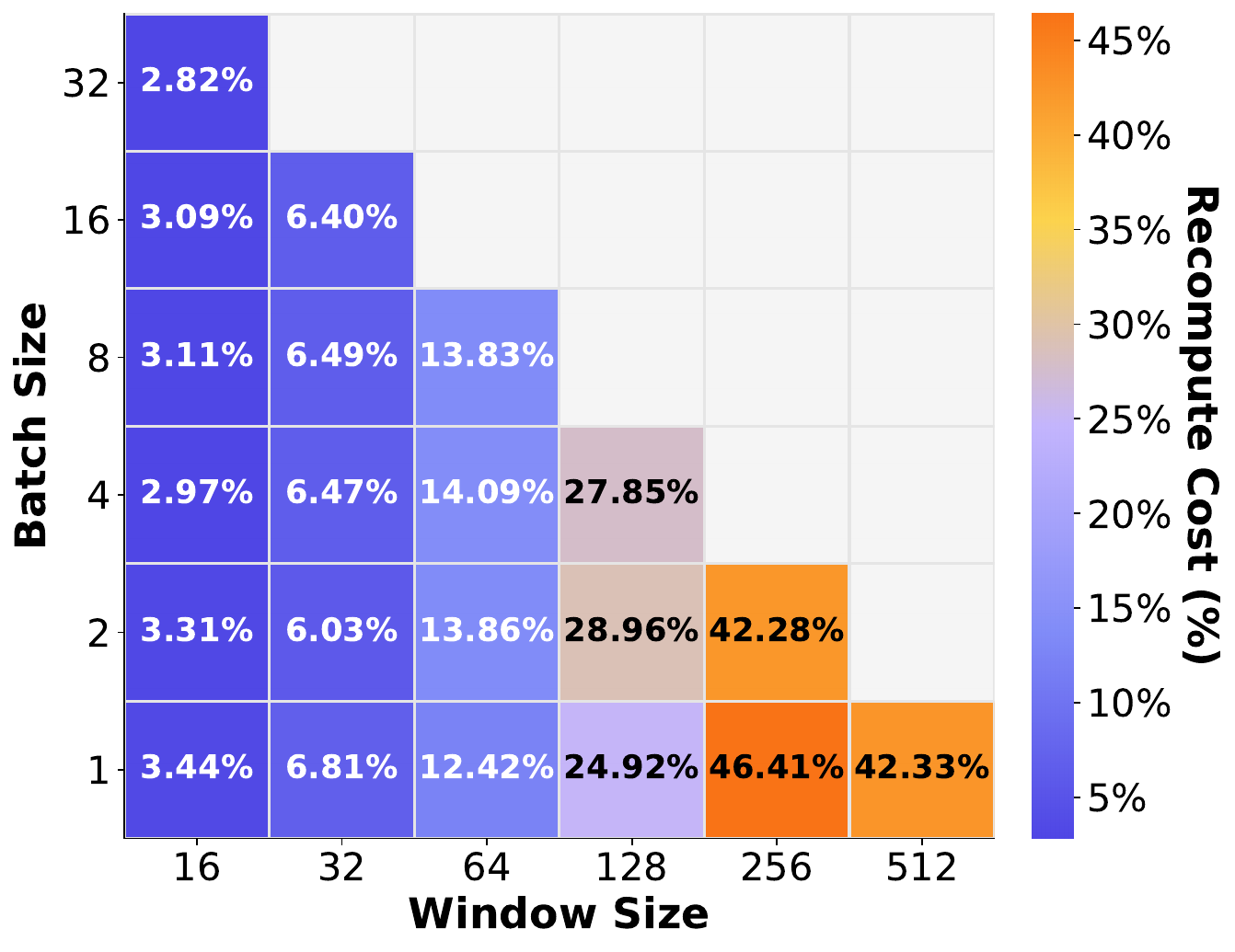}
        \caption{Recompute cost}
        \label{fig:eval:ablation:recomputation}
    \end{subfigure}
    \caption{The effect of different verification strategies on end-to-end latency (left) and recompute cost (right). Batch size denotes the number of requests verified together.}
    \label{fig:eval:ablation}
\end{figure}

Grouped verification helps \sysname reduce both verification and recomputation cost. It is parameterized by (1) the verification window size per request and (2) the number of requests verified together in a singe pass. To isolate the impact of each parameter, we run an ablation on the ShareGPT dataset, varying the per-request window size from 16 to 512 tokens and the number of requests verified together from one to 32. For each configuration, we execute 4096 requests in an online setting at 12 QPS; all requests require determinism. \autoref{fig:eval:ablation} summarizes the results: \autoref{fig:eval:ablation:latency} reports P99 request latency, while \autoref{fig:eval:ablation:recomputation} shows recomputation overhead.

Without grouped verification (batch size 1, first row), latency exhibits a non-monotonic dependence on window size. Increasing the window initially reduces latency, but latency rises when window size becomes too large. Concretely, P99 latency drops from 615 seconds at a window size of 16 to 56.18 seconds at 128, then increases to 99.94 seconds at 512. This behavior reflects the fundamental trade-off between verification and recomputation: smaller windows incur higher verification overhead, while larger windows amplify recomputation cost—for example, recomputation cost is 42.33\% at window size 512, compared to only 3.44\% at window size 16. Without grouped verification, the best performance occurs at window size of 128 where the latency is 56.18 seconds.

Grouped verification substantially lowers the end-to-end latency. Even batching just two requests reduces P99 latency to 43.06 seconds (best case in the second row from the bottom). Increasing the batch size yields further gains, with the best overall configurations verifying a total of 256 tokens per step—distributed across 16, 8, or 4 requests—all achieving P99 tail latencies of 34–35 seconds.

It is worth noting that the recomputation cost in the offline experiments (typically below 2\% as shown in \autoref{tab:eval:rollback_recompute_ws_64_bs_8}) is lower than in the online inference setting (typically more than 6\% as shown in \autoref{fig:eval:ablation:recomputation}). This difference arises because in offline inference, all requests are available at the beginning and the system tries to maximally utilize memory capacity, resulting in more stable batch sizes. In contrast, online inference experiences significantly greater batch-size variability due to fluctuating load, request arrival and departure times; naturally, higher recomputation rates are correlated with this increased batch-size variability.

\subsection*{Discussion}

\running{Evaluation across models and multiple GPUs}
While we report performance results only for \llama in this paper, we have evaluated the correctness of our approach across multiple additional models, including Qwen-4B-Instruct-2507, Qwen3-14B and Qwen3-30B-A3B-Instruct-2507, and across 1-4 GPUs with tensor-parallelism. Our current experimental setup does not support multimem/NVLS and is limited to pairwise NVLink connectivity; consequently, fairly evaluating multi-GPU performance would require a different platform, which we leave to future work. Supporting these models required no additional code changes in \sysname which highlights the simplicity of our approach.

\running{Limitations} In our current prototype, the verification pass introduces latency overhead for all requests, even when determinism is enabled selectively. This overhead could be mitigated by confining verification to a subset of GPU SMs or by deferring verification batches—for example, by prioritizing regular prefills and decodes over verification. A second limitation is that prefill and decode use different reduction strategies in \sysname, making the system non–prefill-decode invariant. As a result, \sysname currently does not support sharing prefix caches across multiple turns of the same request or sharing across requests. Our implementation also does not currently integrate with speculative decoding–based LLM inference. Addressing these limitations is interesting future work.

%% file: Sections/6_RelatedWork.tex
\section{Related Works}
\label{sec:relatedworks}

Recent advances in LLM inference systems have focused on optimizing throughput, latency, and resource utilization~\cite{orca,vllmsosp,sarathiserve2024,tetriinfer,distserve2024,vattention2024,dynamollm2024,pod-attn}. Orca~\cite{orca} pioneered continuous batching, enabling low-latency serving by dynamically scheduling requests at the granularity of individual iterations. vLLM~\cite{vllmsosp} introduced PagedAttention, a memory management abstraction that allows higher batch sizes and reuse of KV cache across requests, dramatically improving system throughput. Sarathi-Serve~\cite{sarathiserve2024} further improved the throughput–latency trade-off through chunked prefills to better exploit GPU parallelism. DistServe~\cite{distserve2024} and Splitwise~\cite{patel2023splitwise} proposed disaggregated inference architectures that separate prefill and decode workloads across specialized servers, mitigating interference. Complementary lines of work such as speculative decoding~\cite{leviathan2022fast, chen2023accelerating, mamou2024dynamic} and FlexGen~\cite{flexgen} pursue orthogonal optimizations by reducing the number of autoregressive steps or offloading memory to CPU and SSDs. While all these systems achieve impressive performance gains, they typically assume non-deterministic execution, leaving the trade-offs between determinism and efficiency  unexplored.

Enabling determinism in LLM inference has recently drawn increasing attention. Song et al.~\cite{song2024greedy} argued that such non-determinism can distort evaluation results, calling for reproducibility-aware benchmarking. Yuan et al.~\cite{yuan2025fp32death} further linked reproducibility failures to mixed-precision arithmetic and fused kernel inconsistencies, showing their impact on multi-step reasoning accuracy.  Rainbird AI~\cite{rainbird2025deterministic} emphasized deterministic inference as a requirement for traceability and compliance in enterprise and safety-critical domains. 

Recent work by Zhang et al. proposes tensor-parallel–invariant kernels that eliminate training–inference mismatch by ensuring bitwise deterministic results across different tensor parallel sizes for LLM inference~\cite{zhang2025-deterministic-tp}. In contrast, existing LLM serving systems typically achieve determinism through batch-invariant computation. We examine the performance and engineering costs of this approach and propose an alternative mechanism for enabling determinism in LLM inference. By leveraging properties of GPU kernels together with characteristics of LLM inference workloads, our approach seeks to reconcile reproducibility with efficiency—a design space that remains largely unexplored in prior work.

%% file: Sections/7_Conclusion.tex
\section{Conclusion}
\label{sec:conclusion}

Enabling determinism in LLM inference remains tedious today. Existing systems achieve determinism by enforcing batch-invariant computation, an approach that is cumbersome in practice: it requires rewriting kernels at a time when both hardware and model architectures are evolving rapidly. Moreover, batch-invariant kernels are inherently suboptimal, as they prevent kernels from adapting their parallelism strategies at runtime. We present \sysname, a simpler alternative that enables determinism in LLM inference by repurposing speculative decoding. \sysname minimizes the need to write new kernels and supports selective enforcement of determinism, incurring runtime overhead only for the fraction of traffic that actually requires it.

%% file: references.bib
@misc{TMA_Engine,
	howpublished = {\url{https://research.colfax-intl.com/tutorial-hopper-tma/}},
	title = {{CUTLASS Tutorial: Mastering the NVIDIA® Tensor Memory Accelerator (TMA)}},
year={2024},
}

@misc{flexgen,
      title={FlexGen: High-Throughput Generative Inference of Large Language Models with a Single GPU}, 
      author={Ying Sheng and Lianmin Zheng and Binhang Yuan and Zhuohan Li and Max Ryabinin and Daniel Y. Fu and Zhiqiang Xie and Beidi Chen and Clark Barrett and Joseph E. Gonzalez and Percy Liang and Christopher Ré and Ion Stoica and Ce Zhang},
      year={2023},
      eprint={2303.06865},
      archivePrefix={arXiv},
      primaryClass={cs.LG}
}

@inproceedings {orca,
author = {Gyeong-In Yu and Joo Seong Jeong and Geon-Woo Kim and Soojeong Kim and Byung-Gon Chun},
title = {Orca: A Distributed Serving System for {Transformer-Based} Generative Models},
booktitle = {16th USENIX Symposium on Operating Systems Design and Implementation (OSDI 22)},
year = {2022},
isbn = {978-1-939133-28-1},
address = {Carlsbad, CA},
pages = {521--538},
url = {https://www.usenix.org/conference/osdi22/presentation/yu},
publisher = {USENIX Association},
month = jul,
}

@article{kaplan2020scalinglaws,
  author       = {Jared Kaplan and
                  Sam McCandlish and
                  Tom Henighan and
                  Tom B. Brown and
                  Benjamin Chess and
                  Rewon Child and
                  Scott Gray and
                  Alec Radford and
                  Jeffrey Wu and
                  Dario Amodei},
  title        = {Scaling Laws for Neural Language Models},
  journal      = {CoRR},
  volume       = {abs/2001.08361},
  year         = {2020},
  url          = {https://arxiv.org/abs/2001.08361},
  eprinttype    = {arXiv},
  eprint       = {2001.08361},
  bibsource    = {dblp computer science bibliography, https://dblp.org}
}

@misc{openai2022gpt4techreport,
  author       = {OpenAI},
  title        = {{GPT-4} Technical Report},
  journal      = {CoRR},
  volume       = {abs/2303.08774},
  year         = {2023},
  url          = {https://doi.org/10.48550/arXiv.2303.08774},
  doi          = {10.48550/arXiv.2303.08774},
  eprinttype    = {arXiv},
  eprint       = {2303.08774},
  bibsource    = {dblp computer science bibliography, https://dblp.org}
}

@misc{arxiv,
  title = {ccdv/arxiv-summarization},
  howpublished = {\url{https://huggingface.co/datasets/ccdv/arxiv-summarization}},
  year={2024},
}

@INPROCEEDINGS{patel2023splitwise,
  author={Patel, Pratyush and Choukse, Esha and Zhang, Chaojie and Shah, Aashaka and Goiri, Inigo and Maleki, Saeed and Bianchini, Ricardo},
  booktitle={2024 ACM/IEEE 51st Annual International Symposium on Computer Architecture (ISCA)}, 
  title={Splitwise: Efficient Generative LLM Inference Using Phase Splitting}, 
  year={2024},
  volume={},
  number={},
  pages={118-132},
  doi={10.1109/ISCA59077.2024.00019}}

@misc{huggingfacesharegpt,
  title = {ShareGPT\_Vicuna\_unfiltered},
  howpublished = {\url{https://huggingface.co/datasets/anon8231489123/ShareGPT_Vicuna_unfiltered/resolve/main/ShareGPT_V3_unfiltered_cleaned_split.json}},
}

@inproceedings{vllmsosp,
author = {Kwon, Woosuk and Li, Zhuohan and Zhuang, Siyuan and Sheng, Ying and Zheng, Lianmin and Yu, Cody Hao and Gonzalez, Joseph and Zhang, Hao and Stoica, Ion},
title = {Efficient Memory Management for Large Language Model Serving with PagedAttention},
year = {2023},
isbn = {9798400702297},
publisher = {Association for Computing Machinery},
address = {New York, NY, USA},
url = {https://doi.org/10.1145/3600006.3613165},
doi = {10.1145/3600006.3613165},
booktitle = {Proceedings of the 29th Symposium on Operating Systems Principles},
pages = {611–626},
numpages = {16},
location = {Koblenz, Germany},
series = {SOSP '23}
}

@misc{flashdecoding,
	howpublished = {\url{https://crfm.stanford.edu/2023/10/12/flashdecoding.html}},
        author = {Tri Dao and Daniel Haziza and Francisco Massa and Grigory Sizov},
	title = {{Flash-Decoding for long-context inference}},
year = {2023}
}

@inproceedings {distserve2024,
author = {Yinmin Zhong and Shengyu Liu and Junda Chen and Jianbo Hu and Yibo Zhu and Xuanzhe Liu and Xin Jin and Hao Zhang},
title = {{DistServe}: Disaggregating Prefill and Decoding for Goodput-optimized Large Language Model Serving},
booktitle = {18th USENIX Symposium on Operating Systems Design and Implementation (OSDI 24)},
year = {2024},
isbn = {978-1-939133-40-3},
address = {Santa Clara, CA},
pages = {193--210},
url = {https://www.usenix.org/conference/osdi24/presentation/zhong-yinmin},
publisher = {USENIX Association},
month = jul
}

@misc{sarathi2023,
      title={SARATHI: Efficient LLM Inference by Piggybacking Decodes with Chunked Prefills}, 
      author={Amey Agrawal and Ashish Panwar and Jayashree Mohan and Nipun Kwatra and Bhargav S. Gulavani and Ramachandran Ramjee},
      year={2023},
      eprint={2308.16369},
      archivePrefix={arXiv},
      primaryClass={cs.LG},
    url={https://arxiv.org/abs/2308.16369}
}

@misc{tetriinfer,
      title={Inference without Interference: Disaggregate LLM Inference for Mixed Downstream Workloads}, 
      author={Cunchen Hu and Heyang Huang and Liangliang Xu and Xusheng Chen and Jiang Xu and Shuang Chen and Hao Feng and Chenxi Wang and Sa Wang and Yungang Bao and Ninghui Sun and Yizhou Shan},
      year={2024},
      eprint={2401.11181},
      archivePrefix={arXiv},
      primaryClass={cs.DC},
      url={https://arxiv.org/abs/2401.11181}, 
}

@inproceedings {sarathiserve2024,
author = {Amey Agrawal and Nitin Kedia and Ashish Panwar and Jayashree Mohan and Nipun Kwatra and Bhargav Gulavani and Alexey Tumanov and Ramachandran Ramjee},
title = {Taming {Throughput-Latency} Tradeoff in {LLM} Inference with {Sarathi-Serve}},
booktitle = {18th USENIX Symposium on Operating Systems Design and Implementation (OSDI 24)},
year = {2024},
isbn = {978-1-939133-40-3},
address = {Santa Clara, CA},
pages = {117--134},
url = {https://www.usenix.org/conference/osdi24/presentation/agrawal},
publisher = {USENIX Association},
month = jul
}

@inproceedings{nanoflow2024,
author = {Zhu, Kan and Gao, Yufei and Zhao, Yilong and Zhao, Liangyu and Zuo, Gefei and Gu, Yile and Xie, Dedong and Tang, Tian and Xu, Qinyu and Ye, Zihao and Kamahori, Keisuke and Lin, Chien-Yu and Wang, Ziren and Wang, Stephanie and Krishnamurthy, Arvind and Kasikci, Baris},
title = {NanoFlow: towards optimal large language model serving throughput},
year = {2025},
isbn = {978-1-939133-47-2},
publisher = {USENIX Association},
address = {USA},
booktitle = {Proceedings of the 19th USENIX Conference on Operating Systems Design and Implementation},
articleno = {41},
numpages = {17},
location = {Boston, MA, USA},
series = {OSDI '25}
}

@inproceedings{vattention2024,
author = {Prabhu, Ramya and Nayak, Ajay and Mohan, Jayashree and Ramjee, Ramachandran and Panwar, Ashish},
title = {vAttention: Dynamic Memory Management for Serving LLMs without PagedAttention},
year = {2025},
isbn = {9798400706981},
publisher = {Association for Computing Machinery},
address = {New York, NY, USA},
url = {https://doi.org/10.1145/3669940.3707256},
doi = {10.1145/3669940.3707256},
booktitle = {Proceedings of the 30th ACM International Conference on Architectural Support for Programming Languages and Operating Systems, Volume 1},
pages = {1133–1150},
numpages = {18},
location = {Rotterdam, Netherlands},
series = {ASPLOS '25}
}

@INPROCEEDINGS{dynamollm2024,
  author={Stojkovic, Jovan and Zhang, Chaojie and Goiri, Íñigo and Torrellas, Josep and Choukse, Esha},
  booktitle={2025 IEEE International Symposium on High Performance Computer Architecture (HPCA)}, 
  title={DynamoLLM: Designing LLM Inference Clusters for Performance and Energy Efficiency}, 
  year={2025},
  volume={},
  number={},
  pages={1348-1362},
  doi={10.1109/HPCA61900.2025.00102}}

@inproceedings{arxiv-summarization,
  title = "A Discourse-Aware Attention Model for Abstractive Summarization of Long Documents",
  author = "Cohan, Arman  and
    Dernoncourt, Franck  and
    Kim, Doo Soon  and
    Bui, Trung  and
    Kim, Seokhwan  and
    Chang, Walter  and
    Goharian, Nazli",
  booktitle = "Proceedings of the 2018 Conference of the North {A}merican Chapter of the Association for Computational Linguistics: Human Language Technologies, Volume 2 (Short Papers)",
  month = jun,
  year = "2018",
  address = "New Orleans, Louisiana",
  publisher = "Association for Computational Linguistics",
  url = "https://aclanthology.org/N18-2097",
  doi = "10.18653/v1/N18-2097",
  pages = "615--621",
  
}

@misc{fa-3,
      title={FlashAttention-3: Fast and Accurate Attention with Asynchrony and Low-precision}, 
      author={Jay Shah and Ganesh Bikshandi and Ying Zhang and Vijay Thakkar and Pradeep Ramani and Tri Dao},
      year={2024},
      eprint={2407.08608},
      archivePrefix={arXiv},
      primaryClass={cs.LG},
      url={https://arxiv.org/abs/2407.08608}, 
}

@misc{he2025nondeterminism,
  author = {Horace He and Thinking Machines Lab},
  title = {Defeating Nondeterminism in LLM Inference},
  year = {2025},
  howpublished = {\url{https://thinkingmachines.ai/blog/defeating-nondeterminism-in-llm-inference/}},
}

@misc{Anadkat2025consistent,
  author = {Shyamal Anadkat and Thinking Machines Lab},
  title = {How to make your completions outputs consistent with the new seed parameter},
  year = {2023},
  howpublished = {\url{https://cookbook.openai.com/examples/reproducible_outputs_with_the_seed_parameter#model-level-features-for-consistent-outputs}},
  
}

@misc{SGLangTeam2025,
  author = {Team SGLang},
  title = {Towards Deterministic Inference in SGLang and Reproducible RL Training},
  year = {2025},
  howpublished = {\url{https://lmsys.org/blog/2025-09-22-sglang-deterministic/}},
}

@misc{vllm-batch-invariance-2025,
  author = {Team vLLM},
  title = {Batch Invariance},
  year = {2025},
  howpublished = {\url{https://docs.vllm.ai/en/latest/features/batch_invariance/}},
}

@inproceedings{atil2024nondeterminism,
    title = "Non-Determinism of ``Deterministic'' {LLM} System Settings in Hosted Environments",
    author = "At{\i}l, Berk  and
      Aykent, Sarp  and
      Chittams, Alexa  and
      Fu, Lisheng  and
      Passonneau, Rebecca J.  and
      Radcliffe, Evan  and
      Rajagopal, Guru Rajan  and
      Sloan, Adam  and
      Tudrej, Tomasz  and
      Ture, Ferhan  and
      Wu, Zhe  and
      Xu, Lixinyu  and
      Baldwin, Breck",
    editor = "Akter, Mousumi  and
      Chowdhury, Tahiya  and
      Eger, Steffen  and
      Leiter, Christoph  and
      Opitz, Juri  and
      {\c{C}}ano, Erion",
    booktitle = "Proceedings of the 5th Workshop on Evaluation and Comparison of NLP Systems",
    month = dec,
    year = "2025",
    address = "Mumbai, India",
    publisher = "Association for Computational Linguistics",
    url = "https://aclanthology.org/2025.eval4nlp-1.12/",
    pages = "135--148",
    ISBN = "979-8-89176-305-0"
}

@article{yuan2025fp32death,
  title        = {Give Me FP32 or Give Me Death? Challenges and Solutions for Reproducible Reasoning},
  author       = {Jiayi Yuan and Hao Li and Xinheng Ding and Wenya Xie and Yu-Jhe Li and Wentian Zhao and Kun Wan and Jing Shi and Xia Hu and Zirui Liu},
  journal      = {arXiv preprint arXiv:2506.09501},
  year         = {2025},
  url          = {https://arxiv.org/abs/2506.09501},
}

@inproceedings{song2024greedy,
    title = "The Good, The Bad, and The Greedy: Evaluation of {LLM}s Should Not Ignore Non-Determinism",
    author = "Song, Yifan  and
      Wang, Guoyin  and
      Li, Sujian  and
      Lin, Bill Yuchen",
    editor = "Chiruzzo, Luis  and
      Ritter, Alan  and
      Wang, Lu",
    booktitle = "Proceedings of the 2025 Conference of the Nations of the Americas Chapter of the Association for Computational Linguistics: Human Language Technologies (Volume 1: Long Papers)",
    month = apr,
    year = "2025",
    address = "Albuquerque, New Mexico",
    publisher = "Association for Computational Linguistics",
    url = "https://aclanthology.org/2025.naacl-long.211/",
    doi = "10.18653/v1/2025.naacl-long.211",
    pages = "4195--4206",
    ISBN = "979-8-89176-189-6"
}

@techreport{rainbird2025deterministic,
  title        = {Deterministic Graph-Based Inference for Guardrailing Large Language Models},
  author       = {Rainbird AI},
  institution  = {Rainbird Technologies Ltd.},
  year         = {2025},
  month        = mar,
  url          = {https://rainbird.ai/wp-content/uploads/2025/03/Deterministic-Graph-Based-Inference-for-Guardrailing-Large-Language-Models.pdf},
  note         = {Accessed: 2025-10-28}
}

@misc{Charlie2025,
  author = {Charlie Parker},
  title = {How can I ensure deterministic text generation with vLLM, and does it support a global set\_seed?},
  year = {2025},
  howpublished = {\url{https://stackoverflow.com/questions/79467847/how-can-i-ensure-deterministic-text-generation-with-vllm-and-does-it-support-a}},
}

@inproceedings{leviathan2022fast,
  title        = {Fast Inference from Transformers via Speculative Decoding},
  author       = {Yaniv Leviathan and Matan Kalman and Yossi Matias},
  booktitle    = {Proceedings of the 40th International Conference on Machine Learning (ICML)},
  year         = {2022},
  url          = {https://arxiv.org/abs/2211.17192}
}

@article{chen2023accelerating,
  title        = {Accelerating Large Language Model Decoding with Speculative Sampling},
  author       = {Charlie Chen and Sebastian Borgeaud and Geoffrey Irving and Jean-Baptiste Lespiau and Laurent Sifre and John Jumper},
  journal      = {arXiv preprint arXiv:2302.01318},
  year         = {2023},
  url          = {https://arxiv.org/abs/2302.01318}
}

@inproceedings{mamou2024dynamic,
  title        = {Dynamic Speculation Lookahead Accelerates Speculative Decoding of Large Language Models},
  author       = {Jonathan Mamou and Oren Pereg and Daniel Korat and Moshe Berchansky and Nadav Timor and Moshe Wasserblat and Roy Schwartz},
  booktitle    = {Proceedings of The 4th NeurIPS Efficient Natural Language and Speech Processing Workshop (PMLR 262)},
  pages        = {456--467},
  year         = {2024},
  url          = {https://proceedings.mlr.press/v262/mamou24a.html}
}

@misc{gond2025tokenweave,
      title={TokenWeave: Efficient Compute-Communication Overlap for Distributed LLM Inference}, 
      author={Raja Gond and Nipun Kwatra and Ramachandran Ramjee},
      year={2025},
      eprint={2505.11329},
      archivePrefix={arXiv},
      primaryClass={cs.DC},
      url={https://arxiv.org/abs/2505.11329}, 
}

@article{integrating_randomness_llm2024,
  title     = {Integrating Randomness in Large Language Models},
  author    = {Zhang, Wei and Li, Rui and Chen, Hao},
  year      = {2024},
  journal   = {arXiv preprint arXiv:2407.03582},
  url       = {https://arxiv.org/abs/2407.03582}
}

@article{diversity_bias_llm2025,
  title     = {A Comprehensive Analysis of Large Language Model Outputs: Similarity, Diversity, and Bias},
  author    = {Kim, Sungwoo and Patel, Divya and Xu, Tian},
  year      = {2025},
  journal   = {arXiv preprint arXiv:2505.09056},
  url       = {https://arxiv.org/abs/2505.09056}
}

@article{tmops2025,
  author = {Thinking Machines},
  title = {Batch Invariant Ops},
  year = {2025},
  howpublished = {https://github.com/thinking-machines-lab/batch_invariant_ops/tree/main},
}

@misc{vllm-batch-invariant-2025,
  author = {Bram Wasti},
  title = {Deepseek-v3 Batch Invariant on 8xH100},
  year = {2025},
  howpublished = {\url{https://github.com/vllm-project/vllm/pull/26609}},
}

@inproceedings{pod-attn,
author = {Kamath, Aditya K. and Prabhu, Ramya and Mohan, Jayashree and Peter, Simon and Ramjee, Ramachandran and Panwar, Ashish},
title = {POD-Attention: Unlocking Full Prefill-Decode Overlap for Faster LLM Inference},
year = {2025},
isbn = {9798400710797},
publisher = {Association for Computing Machinery},
address = {New York, NY, USA},
url = {https://doi.org/10.1145/3676641.3715996},
booktitle = {Proceedings of the 30th ACM International Conference on Architectural Support for Programming Languages and Operating Systems, Volume 2},
pages = {897–912},
numpages = {16}
}

@misc{tritonfusedkernel-splitk-meta,
      title={Accelerating a Triton Fused Kernel for W4A16 Quantized Inference with SplitK work decomposition}, 
      author={Adnan Hoque and Less Wright and Chih-Chieh Yang and Mudhakar Srivatsa and Raghu Ganti},
      year={2024},
      eprint={2402.00025},
      archivePrefix={arXiv},
      primaryClass={cs.DC},
      url={https://arxiv.org/abs/2402.00025}, 
}

@misc{nvidia_cutlass_blog,
  title        = {CUTLASS: Fast Linear Algebra in CUDA},
  author       = {{NVIDIA}},
  howpublished = {\url{https://developer.nvidia.com/blog/cutlass-linear-algebra-cuda/}},
  year         = {2019},
  note         = {NVIDIA Developer Blog}
}

@inproceedings{specdecoding-icml2023,
author = {Leviathan, Yaniv and Kalman, Matan and Matias, Yossi},
title = {Fast inference from transformers via speculative decoding},
year = {2023},
publisher = {JMLR.org},
booktitle = {Proceedings of the 40th International Conference on Machine Learning},
articleno = {795},
numpages = {13},
location = {Honolulu, Hawaii, USA},
series = {ICML'23}
}

@misc{
xia2023speculative,
title={Speculative Decoding: Lossless Speedup of Autoregressive Translation},
author={Heming Xia and Tao Ge and Si-Qing Chen and Furu Wei and Zhifang Sui},
year={2023},
url={https://openreview.net/forum?id=H-VlwsYvVi}
}

@inproceedings{specinfer-2024,
author = {Miao, Xupeng and Oliaro, Gabriele and Zhang, Zhihao and Cheng, Xinhao and Wang, Zeyu and Zhang, Zhengxin and Wong, Rae Ying Yee and Zhu, Alan and Yang, Lijie and Shi, Xiaoxiang and Shi, Chunan and Chen, Zhuoming and Arfeen, Daiyaan and Abhyankar, Reyna and Jia, Zhihao},
title = {SpecInfer: Accelerating Large Language Model Serving with Tree-based Speculative Inference and Verification},
year = {2024},
isbn = {9798400703867},
publisher = {Association for Computing Machinery},
address = {New York, NY, USA},
url = {https://doi.org/10.1145/3620666.3651335},
doi = {10.1145/3620666.3651335},
booktitle = {Proceedings of the 29th ACM International Conference on Architectural Support for Programming Languages and Operating Systems, Volume 3},
pages = {932–949},
numpages = {18},
location = {La Jolla, CA, USA},
series = {ASPLOS '24}
}

@inproceedings{fu2023lookahead,
author = {Fu, Yichao and Bailis, Peter and Stoica, Ion and Zhang, Hao},
title = {Break the sequential dependency of LLM inference using LOOKAHEAD DECODING},
year = {2024},
publisher = {JMLR.org},
booktitle = {Proceedings of the 41st International Conference on Machine Learning},
articleno = {561},
numpages = {20},
location = {Vienna, Austria},
series = {ICML'24}
}

@inproceedings{medusa,
author = {Cai, Tianle and Li, Yuhong and Geng, Zhengyang and Peng, Hongwu and Lee, Jason D. and Chen, Deming and Dao, Tri},
title = {MEDUSA: Simple LLM inference acceleration framework with multiple decoding heads},
year = {2024},
publisher = {JMLR.org},
booktitle = {Proceedings of the 41st International Conference on Machine Learning},
articleno = {203},
numpages = {27},
location = {Vienna, Austria},
series = {ICML'24}
}

@misc{zhang2025-deterministic-tp,
      title={Deterministic Inference across Tensor Parallel Sizes That Eliminates Training-Inference Mismatch}, 
      author={Ziyang Zhang and Xinheng Ding and Jiayi Yuan and Rixin Liu and Huizi Mao and Jiarong Xing and Zirui Liu},
      year={2025},
      eprint={2511.17826},
      archivePrefix={arXiv},
      primaryClass={cs.LG},
      url={https://arxiv.org/abs/2511.17826}, 
}

@misc{nvls-deterministic,
title={Same data all reduce on H20, but results are different},
author={vLLM},
year={2025},
url={https://github.com/NVIDIA/nccl/issues/1497#issuecomment-3210819243}
}

@misc{det-infer-batch-inv-ops-sglang,
title={Support deterministic inference with Batch Invariant Ops},
author={sglang},
year={2025},
url={https://github.com/sgl-project/sglang/issues/10278}
}

@misc{batch-inv-inf-vllm,
title={Batch-invariant Inference},
author={vllm},
year={2025},
url={https://github.com/orgs/vllm-project/projects/29}
}

@misc{det-inf-kills,
      title={Stochastic CHAOS: Why Deterministic Inference Kills, and Distributional Variability Is the Heartbeat of Artifical Cognition}, 
      author={Tanmay Joshi and Shourya Aggarwal and Anusa Saha and Aadi Pandey and Shreyash Dhoot and Vighnesh Rai and Raxit Goswami and Aman Chadha and Vinija Jain and Amitava Das},
      year={2026},
      eprint={2601.07239},
      archivePrefix={arXiv},
      primaryClass={cs.AI},
      url={https://arxiv.org/abs/2601.07239}, 
}
